\documentclass[10pt,twocolumn,letterpaper]{article}

\usepackage{wacv}
\usepackage{times}
\usepackage{epsfig}
\usepackage{graphicx}
\usepackage{amsmath}
\usepackage{amssymb}
\usepackage{subcaption}
\usepackage{caption}
\usepackage{multirow}
\graphicspath{{./figures/}}
\usepackage{arydshln}
\usepackage{color}
\usepackage{cite}
\usepackage{animate}
\usepackage{gensymb}
\usepackage{xurl}

\newcommand{\lossfun}{\mathcal{L}}

\def\wacvPaperID{0462} 

\wacvfinalcopy 

\ifwacvfinal
\def\assignedStartPage{1} 
\fi

\ifwacvfinal
\usepackage[breaklinks=true,bookmarks=false]{hyperref}
\else
\usepackage[pagebackref=true,breaklinks=true,colorlinks,bookmarks=false]{hyperref}
\fi

\ifwacvfinal
\else
\fi

\begin{document}

\title{Improving Single-Image Defocus Deblurring: How Dual-Pixel Images Help Through Multi-Task Learning}
\author{Abdullah Abuolaim \qquad Mahmoud Afifi \qquad Michael S. Brown\\
York University\\
{\tt\small \{abuolaim,mafifi,mbrown\}@eecs.yorku.ca}
}

\maketitle

\begin{abstract}
Many camera sensors use a dual-pixel (DP) design that operates as a rudimentary light field providing two sub-aperture views of a scene in a single capture.  The DP sensor was developed to improve how cameras perform autofocus.  Since the DP sensor's introduction, researchers have found additional uses for the DP data, such as depth estimation, reflection removal, and defocus deblurring.  We are interested in the latter task of defocus deblurring.  In particular, we propose a single-image deblurring network that incorporates the two sub-aperture views into a multi-task framework. Specifically, we show that jointly learning to predict the two DP views from a single blurry input image improves the network's ability to learn to deblur the image. Our experiments show this multi-task strategy achieves $+1$dB PSNR improvement over state-of-the-art defocus deblurring methods.
In addition, our multi-task framework allows accurate DP-view synthesis (e.g., $\sim$ $39$dB PSNR) from the single input image.  These high-quality DP views can be used for other DP-based applications, such as reflection removal. As part of this effort, we have captured a new dataset of $7,059$ high-quality images to support our training for the DP-view synthesis task. Our dataset, code, and trained models are publicly available at  \url{https://github.com/Abdullah-Abuolaim/multi-task-defocus-deblurring-dual-pixel-nimat}.
\end{abstract}

\begin{figure}
     \centering
     \begin{subfigure}[b]{0.15\textwidth}
         \centering
         \includegraphics[width=\textwidth]{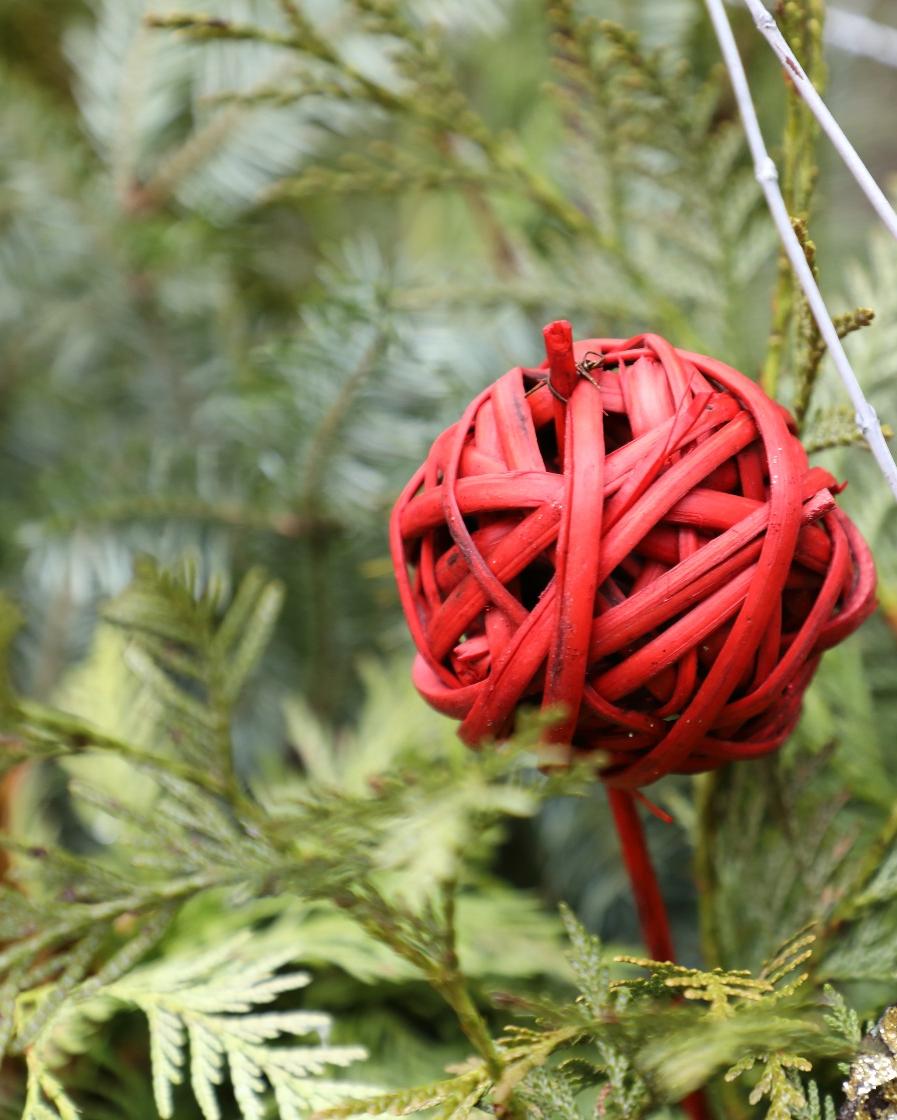}\vspace{-2mm}
         \caption{Blurred input}
         \label{fig:teaser_input}
     \end{subfigure}
    \begin{subfigure}[b]{0.15\textwidth}
          \centering
          \includegraphics[width=\textwidth]{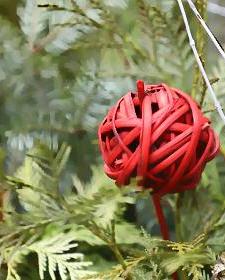}\vspace{-2mm}
          \caption{Result of~\cite{karaali2017edge}}
          \label{fig:loss_C_gt_1}
	\end{subfigure}
     \begin{subfigure}[b]{0.15\textwidth}
         \centering
         \includegraphics[width=\textwidth]{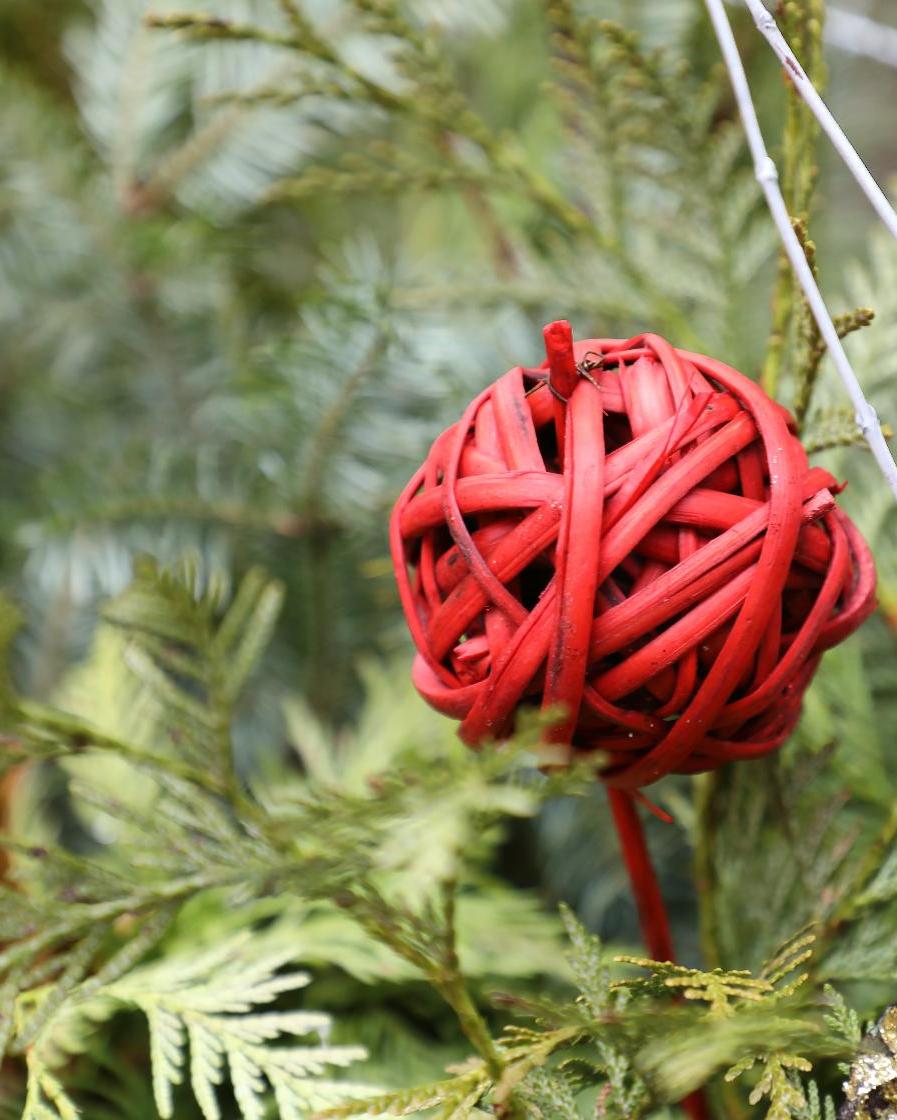}\vspace{-2mm}
         \caption{Result of~\cite{lee2019deep}}
         \label{fig:with_loss_C_1}
     \end{subfigure}
    
    \vspace{1mm}
    \begin{subfigure}[b]{0.15\textwidth}
         \centering
         \includegraphics[width=\textwidth]{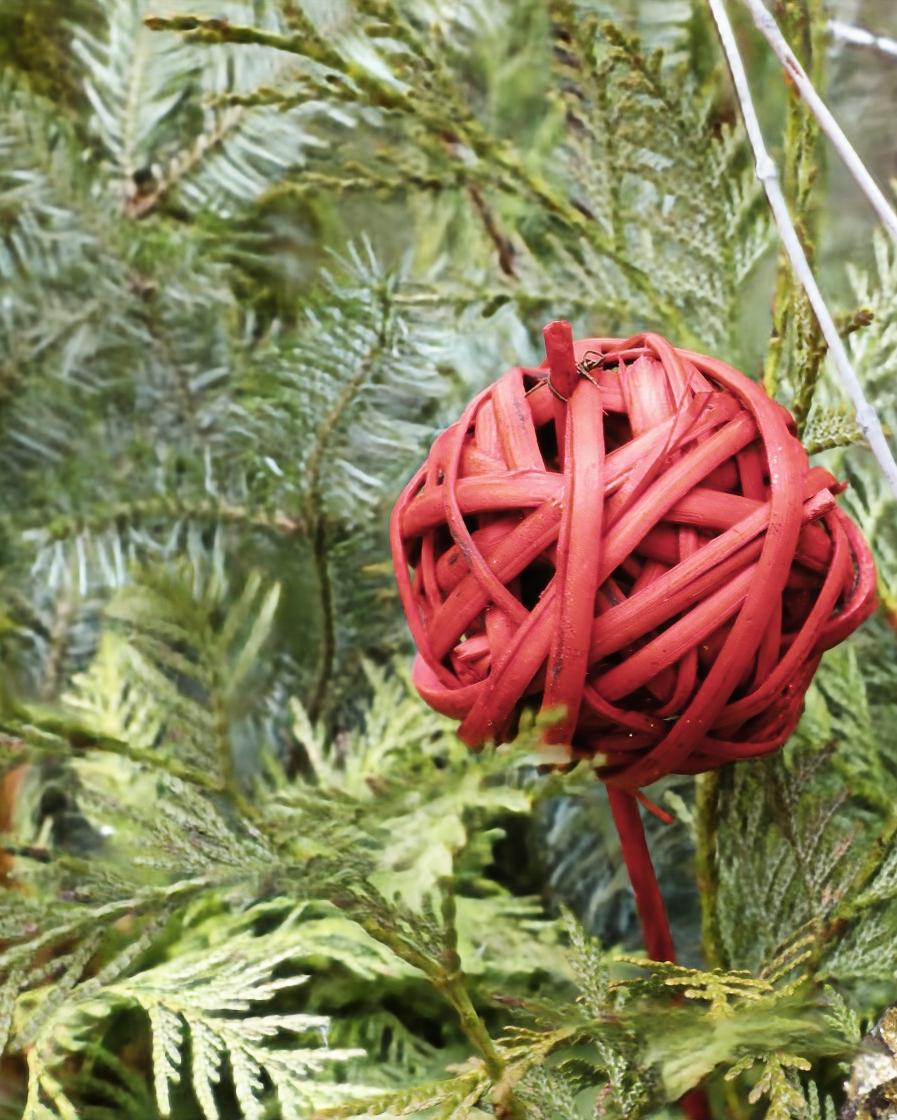}\vspace{-2mm}
         \caption{Our result}
         \label{fig:teaser_ours}
     \end{subfigure}
     \begin{subfigure}[b]{0.15\textwidth}
         \centering
         \animategraphics[width=\textwidth]{3}{teaser_our_dp_views/Slide}{1}{12}\vspace{-2mm}
         \caption{Our DP views}
         \label{fig:teaser_our_dp}
     \end{subfigure}
    \begin{subfigure}[b]{0.15\textwidth}
          \centering
          \animategraphics[width=\textwidth]{3}{teaser_gt_dp_views/Slide}{1}{12}\vspace{-2mm}
          \caption{GT DP views}
          \label{fig:teaser_gt_dp}
	\end{subfigure}
	\vspace{-2mm}
    \caption{Results from our multi-task framework. (a) Input image with DoF blur. Deblurring results of~\cite{karaali2017edge} and~\cite{lee2019deep} are shown in (b) and (c), respectively. (d) Our result. (e) Our reconstructed DP views. (f) Ground-truth DP views. Our multi-task method has better deblurring results and is able to produce accurate DP views from a {\it single-image} input. \textbf{Note: The DP views are animated; click on the image to start the animation. It is recommended to open this PDF in Adobe Acrobat Reader to work properly. This feature is applicable where ``animations" appears in the subsequent figures.}
    }
    \vspace{-4mm}
    \label{fig:teaser_all}
\end{figure}

\section{Introduction and related work}\label{sec:intro}

We are interested in reducing the defocus blur present in captured images.  Defocus blur occurs at scene points that are captured outside a camera's depth of field (DoF). Reducing defocus blur is challenging due to the nature of the spatially varying point spread functions (PSFs) that vary with scene depth~\cite{levin2011understanding, tang2012utilizing}. Most of the existing DoF blur reduction methods~\cite{d2016non,karaali2017edge,lee2019deep,park2017unified,shi2015just} approach the problem in two stages: (1) estimate the defocus map of the input and (2) apply off-the-shelf non-blind deconvolution (e.g., \cite{fish1995blind,krishnan2009fast}) guided by the estimated defocus map.  The performance of these methods is bounded by the DoF map estimation and the effectiveness of the non-blind deconvolution. Additionally, due to the two-stage approach, these methods have a long processing time.

Recent work in~\cite{abuolaim2020defocus} proposed a DoF deblurring method that leveraged the availability of dual-pixel (DP) sensor data.  This method trained a deep neural network (DNN) that uses the DP sensor's two sub-aperture views as input to predict a single deblurred image.  The effectiveness of the method by~\cite{abuolaim2020defocus} is attributed to the DNN's ability to learn the amount of spatially varying defocus blur from the two DP views. This idea stems from the way the DP sensors work. DP sensors were developed as a means to improve the camera's autofocus system. The DP design produces two sub-aperture views of the scene that exhibit differences in phase that are correlated to the amount of defocus blur as shown in Fig.~\ref{fig:teaser_all}-f (details in Sec.\ \ref{sec:DP_image_formation}). A camera adjusts the lens position to minimize phase differences in the two DP views, resulting in a final in-focus image. Researchers have been quick to leverage the DP sub-images for tasks beyond autofocus~\cite{abuolaim2018revisiting,abuolaim2020online}, including depth map estimation~\cite{punnappurath2020modeling,garg2019learning,zhang20202}, defocus deblurring~\cite{abuolaim2021ntire,vo2021attention,lee2021iterative,pan2021dual}, reflection removal~\cite{punnappurath2019reflection}, and synthetic DoF~\cite{wadhwa2018synthetic}.
 
One notable drawback of using DP data is that most cameras do not provide easy access to the DP sensor's two sub-aperture views.  Although DP sensors are used by many cameras, only two cameras currently provide access to DP images (i.e., Canon 5D DSLR camera~\cite{abuolaim2020defocus,punnappurath2020modeling,punnappurath2019reflection} and Google Pixel series smartphone camera~\cite{wadhwa2018synthetic,garg2019learning,zhang20202}).  Even for these devices accessing the DP data has caveats. For example, the Canon 5D requires special software to extract the two views from a saved RAW image. The Google Pixel device requires a special binary and provides DP data only for the green channel of the RAW image. These limitations make the use of DP data at inference time impractical.

In the context of defocus deblurring, training data is required in the form of paired images---one sharp and one blurred.  Training images are obtained by placing a camera on a tripod and capturing an image using a wide aperture (i.e., blurred image with shallow DoF), followed by a second image captured using a narrow aperture (i.e., target sharp image with large DoF).  Care must be taken to ensure that the camera is not moved between aperture adjustments and that the scene content remains stationary.  Such data acquisition is a time-consuming process and does not facilitate collecting larger datasets---for instance, the recent DP defocus deblurring dataset~\cite{abuolaim2020defocus} contains only $500$ of such pairs.

The aforementioned drawbacks of accessing DP data at inference time and the challenges in capturing blurry/sharp paired data for training serve as the impetus for our proposed multi-task learning framework.  In particular, our method focuses on conventional single-image input.  And, while we cannot remove the need for blurred/sharp training data entirely, we demonstrate that the performance of defocus deblurring is improved by incorporating the joint training of predicting the DP views. The training of the DP-view reconstruction task requires only the capture of unpaired DP images in an unrestricted manner with minimal effort. Because we only need access to DP information at training time, inference time becomes much more practical.

\paragraph{Contributions} We introduce a multi-task DNN framework to jointly learn single-image defocus deblurring and DP-based view prediction as shown in Fig.~\ref{fig:teaser_all}. We show that training a DNN to both deblur the image and predict the two sub-aperture DP views improves deblurring results by up to $+1$dB PSNR over existing state-of-the-art methods.  To facilitate this effort, we have captured an unpaired dataset with varying DoF blur consisting of $2,353$ high-quality full-frame images using a DP camera.  This gives a total of $7,059$ images---$2,353$ conventional images and their corresponding two sub-aperture DP views.  We also introduce novel loss functions based on DP image formation to help the network avoid ambiguity that arises in DP data.  Extensive experiments show our results outperform existing single-image DoF deblurring both quantitatively and qualitatively. While our main goal is defocus deblurring, we conclude by showing how our DNN model has the added advantage of being able to produce high-quality DP views that can be used for tasks such as reflection removal and multi-view synthesis.

\section{Dual-pixel image formation}\label{sec:DP_image_formation}

\begin{figure}
    \centering
    \includegraphics[width=\linewidth]{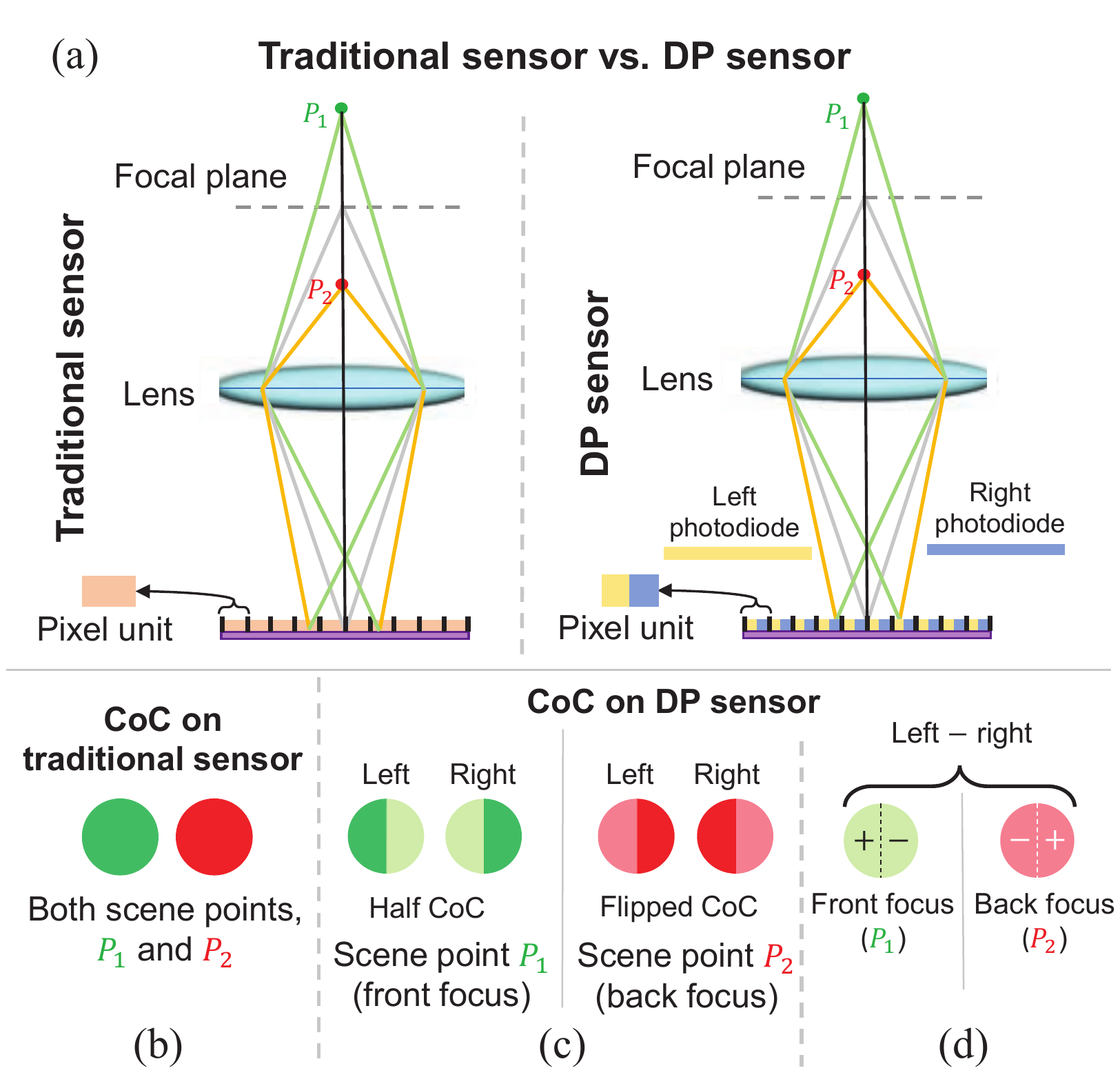}
    \vspace{-2mm}
    \caption{DP sensor image formation via DoF circle of confusion (CoC) formation. (a) Traditional sensor vs. DP sensor. (b) and (c) are the CoC formation on the 2D imaging sensor of two scene points, $P1$ and $P2$. On the two DP views, the half-CoC flips direction if the scene point is in front or back of the focal plane. (d) shows the subtracted DP views in the front/back focus cases, where the $+/-$ sign reveals the front/back focus ambiguity.}
    \vspace{-2mm}
    \label{fig:dp_formation}
\end{figure}

\begin{figure*}[t]
    \centering
    \includegraphics[width=\linewidth]{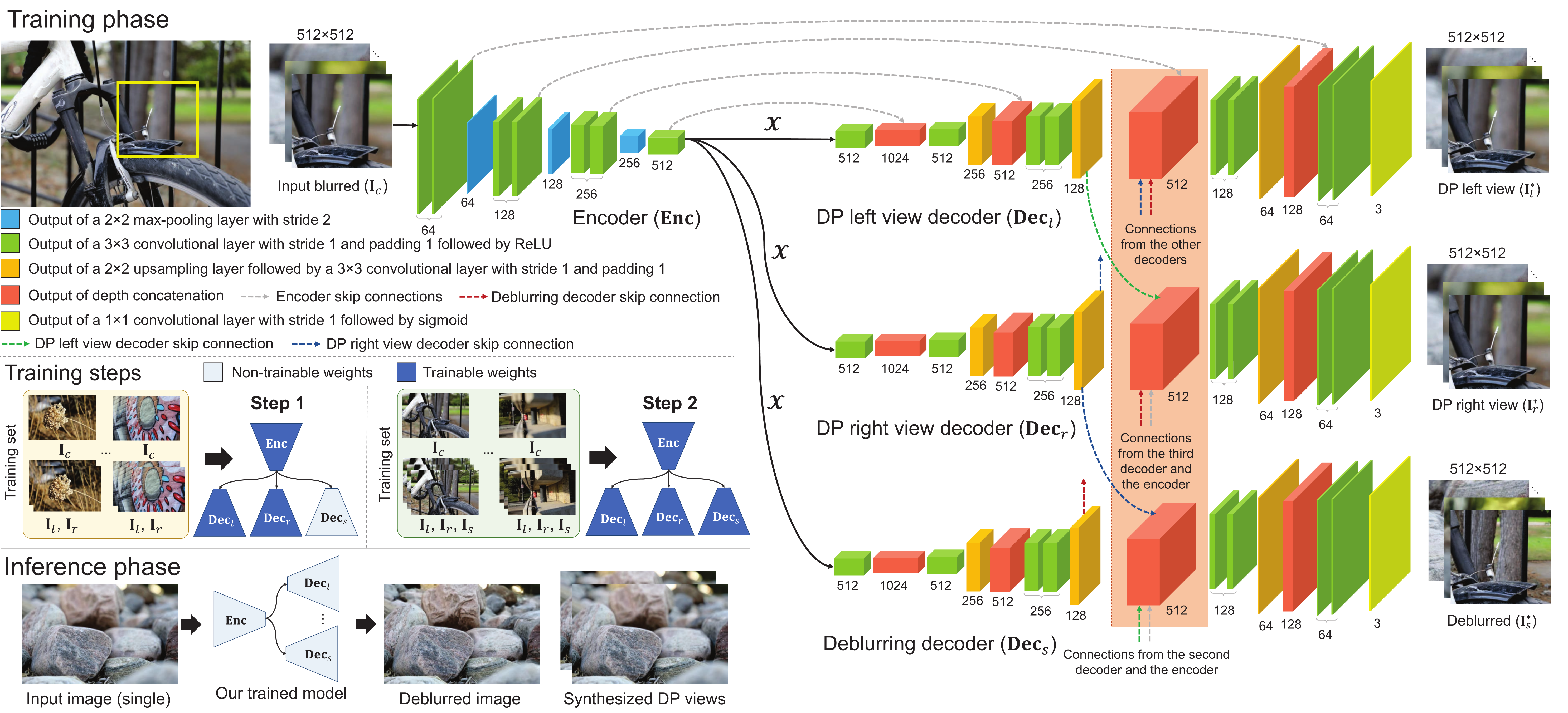}
    \vspace{-4mm}
    \caption{An overview of the proposed multi-task learning framework. We adopt a single-encoder multi-decoder DNN. Our multi-task DP network (MDP) takes a single input image ($\mathbf{I}_c$) and outputs three images---namely, left ($\mathbf{I}_l$) and right ($\mathbf{I}_r$) DP views, and a deblurred (sharp) version ($\mathbf{I}_s$). MDP has two stages of weight sharing between the three decoders (i.e., $\mathbf{Dec}_l$, $\mathbf{Dec}_r$, and $\mathbf{Dec}_s$): early at the encoder ($\mathbf{Enc}_l$) latent space $\mathcal{X}$ and middle at the box highlighted in orange. Our MDP is trained in two steps, where the $\mathbf{Dec}_s$ is frozen in the first step and resumed in the next based on the intended task.}
    \vspace{-2mm}
    \label{fig:overview}
\end{figure*}

We start with a brief overview of DP sensors.  A DP sensor uses two photodiodes at each pixel location with a microlens placed on the top of each pixel site, as shown in Fig.~\ref{fig:dp_formation}-a. As previously mentioned, this design was developed to improve camera autofocus by functioning as a simple two-sample light field camera. The two-sample light field provides two sub-aperture views of the scene and, depending on the sensor's orientation, the views can be referred to as left/right or top/down pairs; we follow the convention of prior papers~\cite{abuolaim2020defocus,punnappurath2020modeling,punnappurath2019reflection} and refer to them as the left/right pair, denoted as $\mathbf{I}_l$ and $\mathbf{I}_r$.   The light rays coming from scene points that are within the camera's DoF exhibit little to no difference in phase between the views. On the other hand, light rays coming from scene points outside the camera's DoF exhibit a noticeable defocus disparity in the {\it left}-{\it right} views. The amount of defocus disparity is correlated to the amount of defocus blur.

Unlike traditional stereo, the difference between the DP views can be modeled as the latent sharp image being blurred in two different directions using a half-circle PSF~\cite{punnappurath2020modeling}. This is illustrated in the resultant circle of confusion (CoC) of Fig.~\ref{fig:dp_formation}-c. On real DP sensors, the ideal case of a half-circle CoC is only an approximation due to constraints of the sensor's construction and lens array. These constraints allow a part of the light ray bundle to leak into the other-half dual pixels (see half CoC of left/right views in Fig.~\ref{fig:dp_formation}-c). We can describe the DP image formation as follows. Let $\mathbf{I}_s$ be a latent sharp image patch and $\mathbf{H}_l$ and $\mathbf{H}_r$ are the left/right PSFs; then the DP $\mathbf{I}_l$ and $\mathbf{I}_r$ can be represented as:
\begin{equation} \label{equ:dpViews}
\mathbf{I}_l = \mathbf{I}_s \ast \mathbf{H}_l, \qquad
\mathbf{I}_r = \mathbf{I}_s \ast \mathbf{H}_r,
\end{equation}
\begin{equation} \label{equ:dpPsfs}
\mathbf{H}_r = \mathbf{H}_l^f,
\end{equation}
where $\ast$ denotes the convolution operation and $\mathbf{H}_l^f$ is the flipped $\mathbf{H}_l$. The two views $\mathbf{I}_l$ and $\mathbf{I}_r$ are combined to produce the final image provided by the camera $\mathbf{I}_c$ as follows:
\begin{equation} \label{equ:dpCombined}
\mathbf{I}_c = \mathbf{I}_l + \mathbf{I}_r.
\end{equation}

Another interesting property of the DP PSFs is that the orientation of the ``half CoC'' of each left/right view reveals if the scene point is in front or back of the focal plane, as shown in the subtracted views of the two scene points, $P_1$ and $P_2$ in Fig.~\ref{fig:dp_formation}-d. These DP properties are useful cues that will be considered when formulating the DP-based loss functions in Sec.~\ref{sec:loss}.

\section{Multi-task learning framework} \label{sec:method}

Multi-task learning has been successfully used for various computer vision tasks~\cite{girshick2015fast,yu2020bdd100k,liu2019end}. In this work, we adopt a multi-task framework in order to leverage the strong connection between defocus deblurring and DP-view synthesis as they require encoding information regarding the defocus blur present at each pixel in the input image. Towards this goal, we propose a single-encoder multi-decoder DNN that takes a single input image and decomposes it into DP left/right views along with the deblurred version. Fig.~\ref{fig:overview} provides an overview of our proposed framework.

\subsection{Network architecture} We adopt a symmetric single-encoder multi-decoder DNN architecture with skip connections between the corresponding feature maps~\cite{mao2016image,ronneberger2015u} (see Fig.~\ref{fig:overview}). We refer to our DNN model as the multi-task DP network (MDP). The three decoder branches have an early-stage weight sharing at the end of the encoder part. We add middle-stage weight sharing as indicated in the orange box of Fig.~\ref{fig:overview}. Each block in the middle-stage sharing receives two skip connections from the corresponding feature maps from the other two decoders. This type of multi-decoder stitching---that guarantees weight sharing at multiple stages---provides multiple communication layers that can further assist the multi-task joint training. We avoid adding late-stage weight sharing as the sharpness of the deblurred image can be affected by the half-PSF blur present in feature maps of the synthesized DP views at these later stages. The proposed model has a sufficiently large receptive field that is able to cover larger spatially varying defocus PSFs. An ablation study is provided in the supplementary material that validates the multi-decoder stitching design.

With the proposed architecture, the encoder $\mathbf{Enc}$ task is to map the input image into a latent space $\mathcal{X}$ as follows:
\begin{equation} \label{equ:encoder}
\mathcal{X} = \mathbf{Enc}(\mathbf{I}_c).
\end{equation}
This latent space can be viewed as a defocus estimation space in which both tasks share a common goal that requires a notion of the PSF size at each pixel in the input image. This latent space representation $\mathcal{X}$ is then passed to the three decoders---namely, left and right DP-view decoders ($\mathbf{Dec}_l$ and $\mathbf{Dec}_r$) and the defocus deblurring (i.e., sharp image) decoder ($\mathbf{Dec}_s$)---in order to produce the output estimations as follows:
\begin{equation} \label{equ:encoder_2}
\mathbf{I}_l^* = \mathbf{Dec}_l(\mathcal{X}), \ \ \ 
\mathbf{I}_r^* = \mathbf{Dec}_r(\mathcal{X}), \ \ \ 
\mathbf{I}_s^* = \mathbf{Dec}_s(\mathcal{X}).
\end{equation}

\subsection{DP-based loss function} \label{sec:loss} It is important to consider how the DP images are formed when designing loss functions to ensure the training process for the two DP views satisfies DP properties.  We have observed empirically that a traditional mean squared error (MSE) loss, computed between the ground truth (GT) and reconstructed DP views, drives the network to a local minima, where the difference between the reconstructed DP views is estimated as an explicit shift in the image content. This observation makes the MSE alone not sufficient to capture the flipping property of DP PSFs (i.e., the PSF reverses direction if it is in front of the focal plane---see Fig.~\ref{fig:dp_formation}-c). Therefore, we introduce a novel DP-loss based on Eq.~\ref{equ:dpCombined} that imposes a constraint on the DP-view reconstruction process as follows:
\begin{equation}\label{equ:dpLoss_1}
    \mathcal{L}_C=\frac{1}{n}\sum_n(\mathbf{I}_c-(\mathbf{I}_l^*+\mathbf{I}_r^*))^2,
\end{equation}
where $\mathbf{I}_c$ is the input combined image and $\mathbf{I}_l^*$ and $\mathbf{I}_r^*$ are the estimated DP views. Our $\mathcal{L}_C$ encourages the network to optimize for the fundamental DP image formation (i.e., Eq.~\ref{equ:dpCombined}).

While $\mathcal{L}_C$ assists the network to learn that the combined left/right views should sum to the combined image, the front/back focus flipping direction remains ambiguous to the network. To address this ambiguity, we introduce a new view difference loss $\mathcal{L}_D$ to capture the flipping sign direction as follows:
\begin{equation}\label{equ:dpLoss_2}
    \mathcal{L}_D=\frac{1}{n}\sum_n((\mathbf{I}_l-\mathbf{I}_r)-(\mathbf{I}_l^*-\mathbf{I}_r^*))^2,
\end{equation}
where $\mathbf{I}_l$ and $\mathbf{I}_r$ are the GT DP left and right views, respectively. Fig.~\ref{fig:dp_formation}-d shows the sign difference in the front/back focus cases when the views are subtracted, which gives a cue for the network to learn the PSF flipping direction when penalizing view difference in the loss---namely, $\mathcal{L}_D$. Fig.\ \ref{fig:loss} provides a visual analysis of the observations that led to our two DP loss functions---namely, $\mathcal{L}_C$ and $\mathcal{L}_D$. See Sec.\ \ref{sec:ablation} for an ablation to demonstrate their effectiveness.

\begin{figure*}[t]
     \centering
     \begin{subfigure}[b]{0.32\textwidth}
         \centering
         \animategraphics[width=\textwidth]{3}{loss0a/Slide}{1}{12}
         \caption{Without $\lossfun_C$ loss term}
         \label{fig:without_loss_C}
     \end{subfigure}
     \begin{subfigure}[b]{0.32\textwidth}
         \centering
         \animategraphics[width=\textwidth]{3}{loss0b/Slide}{1}{12}
         \caption{With $\lossfun_C$ loss term}
         \label{fig:with_loss_C_2}
     \end{subfigure}
    \begin{subfigure}[b]{0.32\textwidth}
          \centering
          \animategraphics[width=\textwidth]{3}{loss0c/Slide}{1}{12}
          \caption{Ground truth}
          \label{fig:loss_C_gt_2}
	\end{subfigure}
    
    \begin{subfigure}[b]{0.32\textwidth}
         \centering
         \animategraphics[width=\textwidth]{3}{loss1a/Slide}{1}{12}
         \caption{Without $\lossfun_D$ loss term}
         \label{fig:without_loss_D}
     \end{subfigure}
     \begin{subfigure}[b]{0.32\textwidth}
         \centering
         \animategraphics[width=\textwidth]{3}{loss1b/Slide}{1}{12}
         \caption{With $\lossfun_D$ loss term}
         \label{fig:with_loss_D}
     \end{subfigure}
    \begin{subfigure}[b]{0.32\textwidth}
          \centering
          \animategraphics[width=\textwidth]{3}{loss1c/Slide}{1}{12}
          \caption{Ground truth}
          \label{fig:loss_D_gt}
	\end{subfigure}
     \vspace{-2mm}
    \caption{ 
    An analysis and visual comparison to reflect the effectiveness of our proposed $\lossfun_C$ and $\lossfun_D$ DP-based loss terms. (b) shows that training with $\lossfun_C$ helps the network to capture the flipping kernel (yellow patch) and accurate colors (red patch) compared to the one in (a). (e) demonstrates that training with $\lossfun_D$ can assist the network to learn the flipping direction in the front (yellow patch) and back focus (red patch), where the views rotate around the focal plane as shown in the GT. {\bf Note: the images are animated; click on the image to start the animation. It is recommended to open this PDF in Adobe Acrobat Reader to work properly.}}
    \label{fig:loss}
    \vspace{-4mm}
\end{figure*}

\subsection{Dual-pixel datasets}
\paragraph{Our new DP dataset} We collected a new \emph{diverse} and \emph{large} DP (DLDP) dataset of $2,353$ scenes. These scenes were captured using a Canon EOS 5D DSLR camera. Each scene consists of a high-quality combined image ($2,353$ images) with its corresponding DP views ($2,353\times2$ images). All images are captured at full-frame resolution (i.e., 6720 × 4480 pixels). Our data contains indoor and outdoor scenes with diverse image content, weather conditions, scene illuminations, and day/night scenes. We captured scenes with different aperture sizes (i.e., $f/4$, $f/5.6$, $f/10$, $f/16$, and $f/22$) in order to cover a wider range of spatially varying defocus blur (i.e., from all-in-focus to severely blurred images). We use this DLDP dataset in our multi-task framework to optimize directly for the DP-view synthesis task.

\paragraph{Other DP datasets} We use the Canon DP deblurring dataset~\cite{abuolaim2020defocus} (i.e., $350$ training paired images) to optimize for both defocus deblurring and DP-view synthesis. We note that there is also a DP dataset based on the Google Pixel camera~\cite{garg2019learning}.  We opted to only use the Canon as it matches our DLDP dataset. Additionally, the Pixel DP data from~\cite{garg2019learning} is not appropriate to train for colored images as it provides only the green channels in the raw-Bayer frame for the DP views.

\subsection{Model training} Our training is divided into two steps. First, training with image patches from our DLDP dataset is performed to optimize only the DP-view synthesis task. During this step the weights of the deblurring decoder branch ($\mathbf{Dec}_s$) are frozen. Once the model converges for the DP-view synthesis branches, the second step unfreezes the weights of $\mathbf{Dec}_s$ and starts fine-tuning using image patches from the Canon DP deblurring dataset~\cite{abuolaim2020defocus} to optimize jointly for both the defocus deblurring and DP-view synthesis tasks (see training steps in Fig.~\ref{fig:overview}). For the first step, we train the network with the following loss terms:
\begin{equation}\label{equ:dpLoss_3}
    \mathcal{L}_{ST1}=\mathcal{L}_{MSE}(l,r)+\mathcal{L}_{C}+\mathcal{L}_{D},
\end{equation}
where $\mathcal{L}_{ST1}$ is the overall first-step loss and $\mathcal{L}_{MSE}(l,r)$ is the typical MSE loss between the GT and estimated DP views. The second step needs more careful loss setting to fine-tune the model in a way that guarantees improving performance on both tasks. In the second step, the network is fine-tuned with the following loss terms:

\begin{equation}\label{equ:dpLoss_4}
     \mathcal{L}_{ST2}= \mathcal{L}_{MSE}(s) + \lambda_{1} \mathcal{L}_{MSE}(l,r) \\
+ \lambda_{2} \mathcal{L}_{C} + \lambda_{3} \mathcal{L}_{D} 
\end{equation}
where $\mathcal{L}_{ST2}$ is the overall second-step loss and $\mathcal{L}_{MSE}(s)$ is the typical MSE between the output deblurred image and the GT. The $\lambda$ terms are added to control the training process.

\section{Experiments} \label{sec:results}

\begin{table*}[t]
\caption{Quantitative comparisons with single-image defocus deblurring methods. The best results are indicated with boldface. Results are on the Canon DP deblurring dataset \cite{abuolaim2020defocus} (test set consists of 37 indoor and 39 outdoor scenes). 
\label{table:single_image_deblurring}}
\vspace{-2mm}
\centering
\scalebox{0.8}{
\begin{tabular}{l|ccc|ccc|ccccc}
\multicolumn{1}{l|}{\multirow{2}{*}{Method}} &  \multicolumn{3}{c|}{Indoor} &  \multicolumn{3}{c|}{Outdoor} &  \multicolumn{5}{c}{Indoor \& Outdoor} \\ \cline{2-12}
\multicolumn{1}{l|}{}  & PSNR $\uparrow$ & SSIM $\uparrow$ & MAE $\downarrow$ & PSNR $\uparrow$ & SSIM $\uparrow$ & MAE $\downarrow$ & PSNR $\uparrow$ & SSIM $\uparrow$ & MAE $\downarrow$ & NIQE $\downarrow$ & Time $\downarrow$\\ \hline
JNB \cite{shi2015just} & 26.73 & 0.828 & 0.031 & 21.10 & 0.608 & 0.064 & 23.84 & 0.715 & 0.048 & 5.11 & 843.1 \\
EBDB \cite{karaali2017edge} &  25.77 & 0.772 &0.040 & 21.25 & 0.599 & 0.058 & 23.45 & 0.683 & 0.049 & 5.42 & 929.7 \\
DMENet \cite{lee2019deep} & 25.70 & 0.789 & 0.036 & 21.51 & 0.655 & 0.061 & 23.55 & 0.720 & 0.049 & 4.85 & 613.7 \\
DPDNet (single) \cite{abuolaim2020defocus} & 26.54 & 0.816 & 0.031 & 22.25 & 0.682 & 0.056  & 24.34 & 0.747 & 0.044 & 4.06 & {\bf 0.5} \\
Our MDP  & {\bf 28.02} & {\bf 0.841} & {\bf 0.027} & {\bf 22.82} & {\bf 0.690} & {\bf 0.052} & {\bf 25.35} & {\bf 0.763} & {\bf 0.040} & {\bf 3.25} &  {\bf 0.5}\\
\end{tabular}}
\vspace{-2mm}
\end{table*}

\begin{table}[t]
\caption{Our single-image defocus deblurring achieves on-par results 
compared to DPDNet \cite{abuolaim2020defocus}, which requires two DP images data as input. Results are on the Canon DP deblurring dataset \cite{abuolaim2020defocus}.\label{table:dp_deblurring}}
\vspace{-2mm}
\centering
\scalebox{0.84}{
\begin{tabular}{l|ccc}
Method & PSNR $\uparrow$ & SSIM $\uparrow$ & MAE $\downarrow$ \\ \hline
DPDNet (real DP views) \cite{abuolaim2020defocus} & 25.13 & {\bf 0.786} & 0.041 \\
DPDNet (our synth. DP views) \cite{abuolaim2020defocus} & 24.91 & 0.758 & 0.043 \\\hdashline
Our MDP (single image) & {\bf 25.35} & 0.763 & {\bf 0.040} \\
\end{tabular}}
\vspace{-2mm}
\end{table}

\subsection{Training details}
\paragraph{Training data}
We divide our newly captured DLDP dataset into $2,090$ and $263$ training and testing scenes, respectively. In the first training step, we use the $2,090$ training scenes. For the second training step, we use the DP data from~\cite{abuolaim2020defocus} following the same data division in~\cite{abuolaim2020defocus}---that is, $350$, $74$, and $76$ training, validation, and testing scenes, respectively.

\paragraph{Training procedure}
We extract image patches of size $512\times512\times3$, where the input is a single patch and the output is three patches. The convolutional layer weights are initialized using He’s method~\cite{he2015delving} and the Adam optimizer~\cite{kingma2014adam} is used to train the model. The mini-batch size in each iteration is set to $8$ batches.

For the first training step, the initial learning rate is set to $3\times10^{-4}$, which is decreased by half every $8$ epochs. The model converges after $60$ epochs in the first step. For the second step, the initial learning rate is set to $6\times10^{-4}$, which is decreased by half every $8$ epochs. The model converges after $80$ epochs. The $\lambda$ terms are set to $0.8$, $0.5$, and $0.5$, respectively, in order to have a balanced loss minimization and to guide the network attention towards minimizing for defocus deblurring in the second step. All the training details and model setup are implemented using Python with the Keras framework on top of TensorFlow 2 and trained with NVIDIA TITAN X GPU. 

\subsection{Single-image deblurring} \label{sec:results_single_image_deblurring}
To evaluate our method, we use the test set of the Canon DP deblurring dataset \cite{abuolaim2020defocus}. Specifically, this test set \cite{abuolaim2020defocus} includes 37 indoor and 39 outdoor scenes. We compare our results against recent methods for single-image defocus deblurring~\cite{shi2015just,karaali2017edge,lee2019deep}. For the sake of completeness, we also compare our results against the recent DP defocus deblurring method (DPDNet)~\cite{abuolaim2020defocus}, which requires the availability of DP data at inference time. As our task is single-image defocus deblurring, we provide the results of the DPDNet \cite{abuolaim2020defocus} using single input images for a fair comparison.
The single-image model of this method was originally trained by the authors of the paper \cite{abuolaim2020defocus} with the same model size (i.e., number of weights) used for the DP model. Although our MDP model has three decoder branches, we adjust the number of convolutional operations and filter size in some blocks in order to have an equivalent model size in terms of number of weights compared to DPDNet~\cite{abuolaim2020defocus}. Our proposed MDP is fully convolutional so that we can test on full-size images regardless of the patch size used for training.

\paragraph{Quantitative results}
Table \ref{table:single_image_deblurring} shows the quantitative results of our method and other single-image defocus deblurring methods: the just noticeable defocus blur method (JNB) \cite{shi2015just}, the edge-based defocus blur estimation method (EBDB) \cite{karaali2017edge}, the deep defocus map estimation method (DMENet) \cite{lee2019deep}, and the DPDNet (single) \cite{abuolaim2020defocus}. We use the common signal processing metrics PSNR, SSIM, and MAE. We also report the Naturalness Image Quality (NIQE) metric of the output deblurred images with respect to a reference model derived from the DP GT images. As shown in Table \ref{table:single_image_deblurring}, our method achieves the state-of-the-art results for all metrics compared to other recent single-image defocus deblurring methods. Furthermore, MDP and DPDNet (single) have much lower inference time---that is, $>$1,200$\times$ faster compared to others. 

Though motion blur leads to image blur too, as defocus blur does, the physical formation and consequently the appearance of the resultant blur are different. This was noted in~\cite{abuolaim2020defocus}, and we found a significant degradation on the accuracy of  methods focused on motion blur~\cite{tao2018scale,zhou2019spatio,nah2019recurrent,su2017deep,pan2020cascaded} when they are applied to defocus blur; for example, Tao et al.'s method \cite{tao2018scale} for motion deblurring achieves an average PSNR of $20.12$dB when it is evaluated on the Canon DP deblurring test set \cite{abuolaim2020defocus}, which is significantly lower than all other defocus deblurring methods shown in Table \ref{table:single_image_deblurring}.

In Table \ref{table:single_image_deblurring}, we reported the results of the DP-based method (i.e., DPDNet) \cite{abuolaim2020defocus} trained on single input. For the sake of completeness, we also compared our method against this method when it is fed with real DP data as input. Table \ref{table:dp_deblurring} shows this comparison. As can be seen, our method achieves higher PSNR and MAE but lower SSIM compared to DPDNet~\cite{abuolaim2020defocus}, while our method is more practical as it  requires only   single-input images compared to the DPDNet~\cite{abuolaim2020defocus}, which requires accessing the two DP images at the inference phase. Recall that DPDNet cannot be trained on our DLDP dataset as it is unpaired and does not have the corresponding GT all-in-focus image. This point is a central strength of our approach, as the use of unpaired DP images enables capturing a much larger dataset for training, thereby making the learning process more efficient.

\begin{figure*}
\newcommand{\wlen}{1}
\centering
     \begin{subfigure}[]{\wlen\textwidth}
         \includegraphics[width=\textwidth]{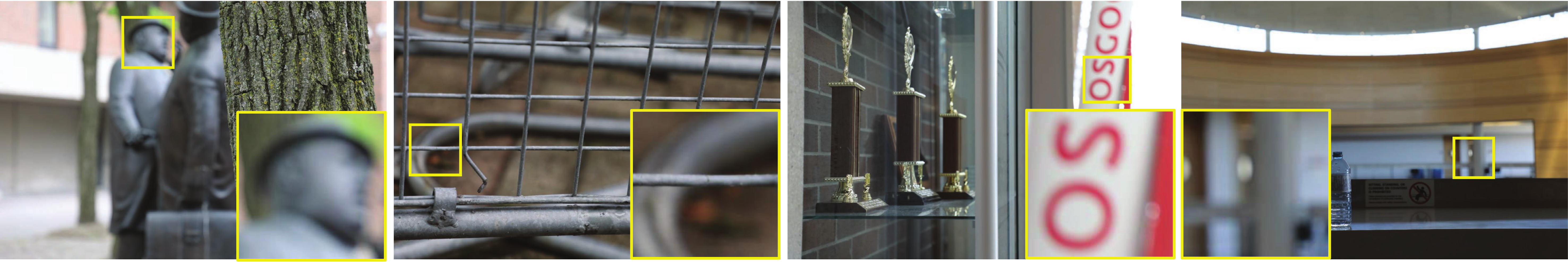}
         \vspace{-6mm}
         \caption{Input}
     \end{subfigure}
     
     \begin{subfigure}[]{\wlen\textwidth}
         \includegraphics[width=\textwidth]{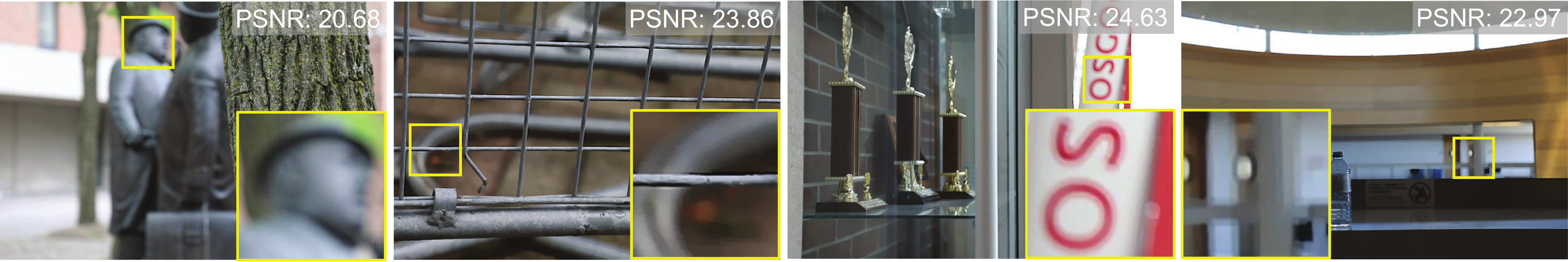}
         \vspace{-6mm}
         \caption{EBDB \cite{karaali2017edge}}
     \end{subfigure}
     
     \begin{subfigure}[]{\wlen\textwidth}
         \includegraphics[width=\textwidth]{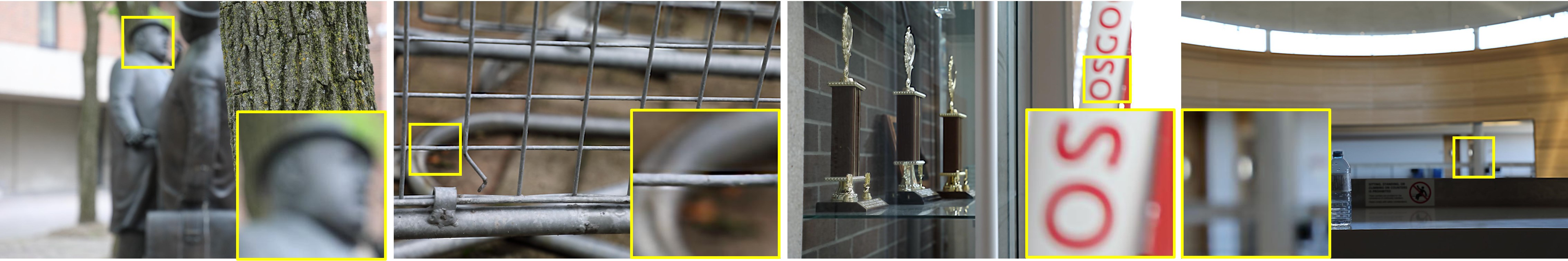}
         \vspace{-6mm}
         \caption{DMENet \cite{lee2019deep}}
     \end{subfigure}
     
     \begin{subfigure}[]{\wlen\textwidth}
         \includegraphics[width=\textwidth]{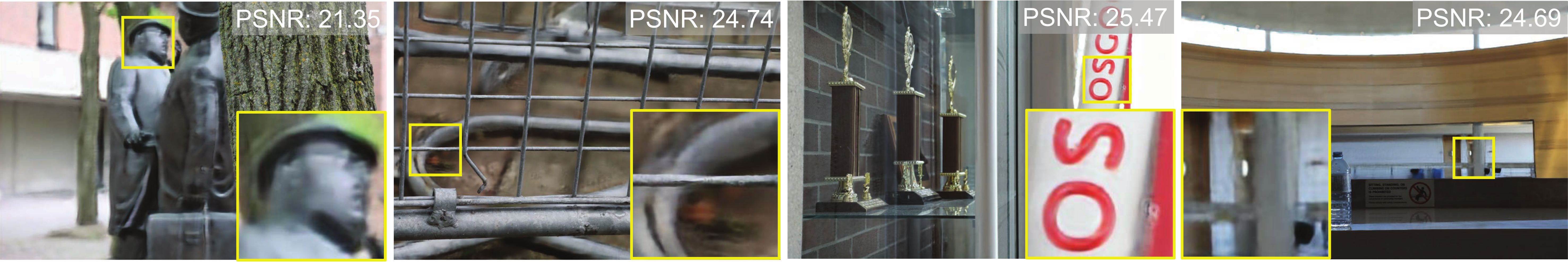}
         \vspace{-6mm}
         \caption{DPDNet (single) \cite{abuolaim2020defocus}}
     \end{subfigure}
     
     \begin{subfigure}[]{\wlen\textwidth}
         \includegraphics[width=\textwidth]{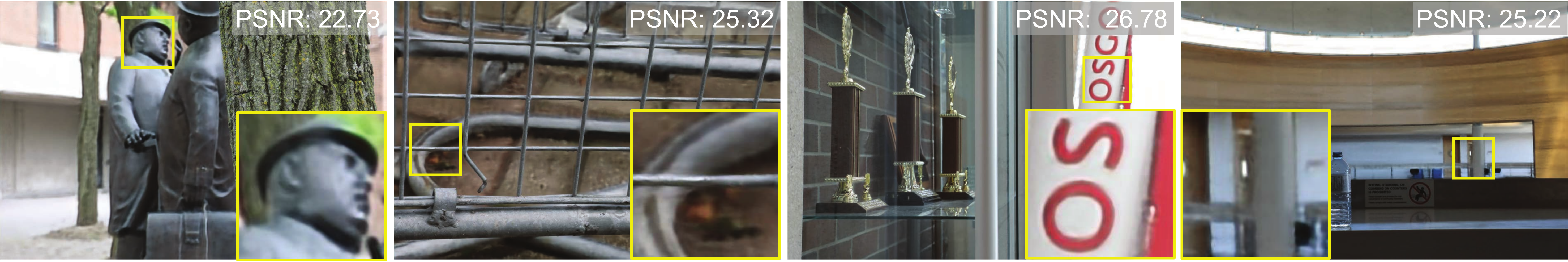}
         \vspace{-6mm}
         \caption{Ours}
     \end{subfigure}
     
     \begin{subfigure}[]{\wlen\textwidth}
         \includegraphics[width=\textwidth]{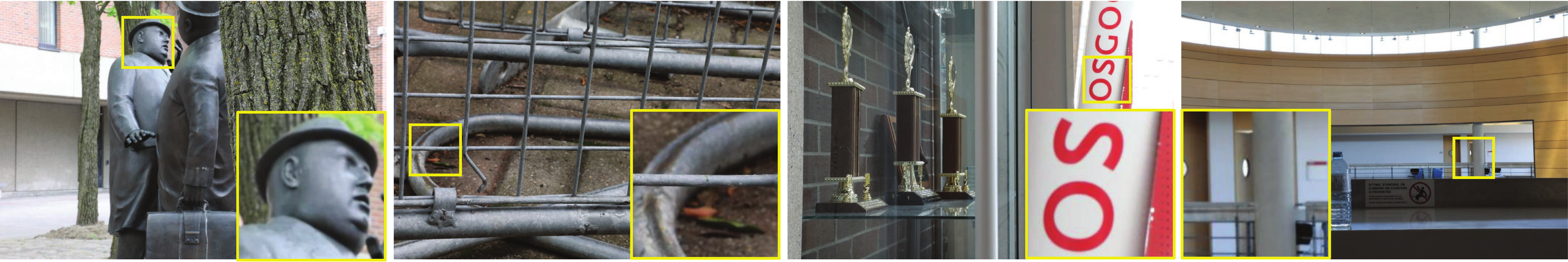}
         \vspace{-6mm}
         \caption{Ground truth}
     \end{subfigure}
    \vspace{-2mm}
    \caption{Qualitative comparisons with other single-image defocus deblurring methods on the test set of the Canon DP dataset~\cite{abuolaim2020defocus}. We compare our results with the following single-image defocus deblurring methods:  EBDB~\cite{karaali2017edge}, DMENet~\cite{lee2019deep}, and DPDNet (single)~\cite{abuolaim2020defocus}. Note that DPDNet was originally introduced to use DP images as input, but the authors in~\cite{abuolaim2020defocus} also provided the same model trained on a single image, denoted as DPDNet (single). Our method produces the best quantitative and arguably qualitative results.}
        \label{fig:qualitative}
    \vspace{-2mm}
\end{figure*}

\paragraph{Qualitative results} 
Our method achieves better qualitative results when compared with several existing single-image defocus deblurring methods (as shown in Fig.\ \ref{fig:teaser_all}). We provide additional qualitative comparisons in Fig.\ \ref{fig:qualitative}, where we compare our results against the results of the EBDB \cite{karaali2017edge}, DMENet \cite{lee2019deep}, and the DPDNet (single) \cite{abuolaim2020defocus} methods. Fig.~\ref{fig:qualitative} shows that our method achieves results that are arguably visually superior to the other methods. Additional results on other cameras are also available in the supplementary material. 

\begin{table}[t]
\caption{Ablation study to demonstrate the effectiveness of our DP-based loss terms and multi-task training. Results are on the Canon DP deblurring dataset \cite{abuolaim2020defocus}. Note that the $\lossfun_C$ and $\lossfun_D$ are not applicable (N/A) to the single task defocus deblurring as it does not predict the DP views.
\label{table:ablation_study}}
\vspace{-2mm}
\centering
\scalebox{0.78}{
\begin{tabular}{l|cc|cc}
\multicolumn{1}{l|}{\multirow{2}{*}{Method}} &  \multicolumn{2}{c|}{Defocus deblurring} &  \multicolumn{2}{c}{DP-pixel synthesis} \\ \cline{2-5}
\multicolumn{1}{l|}{}  & PSNR $\uparrow$ & SSIM $\uparrow$  & PSNR $\uparrow$ & SSIM $\uparrow$   \\ \hline
Single-task w/o $\lossfun_C$ and $\lossfun_D$ & 24.34 & 0.747 & 37.05 & 0.953 \\
Single-task w/ $\lossfun_C$ and $\lossfun_D$ & N/A & N/A & 38.23 & 0.962  \\
Multi-task w/o $\lossfun_C$ and $\lossfun_D$ & 24.81 & 0.750 & 38.01 & 0.957 \\
Multi-task w/ $\lossfun_C$ and $\lossfun_D$ & {\bf 25.35}  & {\bf 0.763} & {\bf 39.17 } & {\bf 0.973} \\ 
\end{tabular}}
\end{table}

\begin{table}[t]
\caption{Reflection removal quantitative results on the dataset proposed in \cite{punnappurath2019reflection}. When using our synthetic DP views, the dual-pixel reflection removal (DPRR) method \cite{punnappurath2019reflection} achieves on-par results compared with using real DP views, which makes the DPRR method applicable with the absence of real DP data. 
\label{table:reflection_removal}}
\vspace{-2mm}
\centering
\scalebox{0.89}{
\begin{tabular}{l|l|ll}
\multicolumn{2}{c|}{Single-image} & \multicolumn{2}{c}{Non-single-image} \\ \hline
Method & PSNR & \multicolumn{1}{l|}{Method} & PSNR \\ \hline
ZN18 \cite{zhang2018single} & 15.57 & \multicolumn{1}{l|}{LB13  \cite{li2013exploiting}} &  16.12 \\
YG18 \cite{yang2018seeing}  & 16.49 & \multicolumn{1}{l|}{GC14 \cite{guo2014robust}} &  16.02 \\
DPRR  \cite{punnappurath2019reflection} (ours) & \textbf{19.32} & \multicolumn{1}{l|}{DPRR \cite{punnappurath2019reflection} (real DP)} & \textbf{19.45}
\end{tabular}}
\vspace{-1mm}
\end{table}

\begin{figure*}[t!]
     \centering
     \begin{subfigure}[b]{0.24\textwidth}
         \centering
         \animategraphics[width=\textwidth]{12}{animation_example_0/000_}{1}{55}
     \end{subfigure}
         \begin{subfigure}[b]{0.24\textwidth}
          \centering
         \animategraphics[width=\textwidth]{12}{animation_example_5/005_}{1}{55}
	\end{subfigure}
     \begin{subfigure}[b]{0.24\textwidth}
         \centering
         \animategraphics[width=\textwidth]{12}{animation_example_3/flickr_0_}{1}{55}
     \end{subfigure}
    \begin{subfigure}[b]{0.24\textwidth}
         \centering
         \animategraphics[width=\textwidth]{12}{animation_example_2/flickr_1_}{1}{55}
     \end{subfigure}
    \vspace{-2mm}
    \caption{ 
    Our novel NIMAT effect. Aesthetically pleasing image motion produced by our multi-view synthesis. 
    The first two input images were taken from the Canon DP deblurring dataset \cite{abuolaim2020defocus}, while the others are from Flickr. {\bf Note: the images are animated; click on the image to start the animation. It is recommended to open this PDF in Adobe Acrobat Reader to work properly.}}
    \label{fig:multi_view} 
    \vspace{-2.5mm}
\end{figure*}

\begin{figure}[t!]
    \centering
    \includegraphics[width=0.94\linewidth]{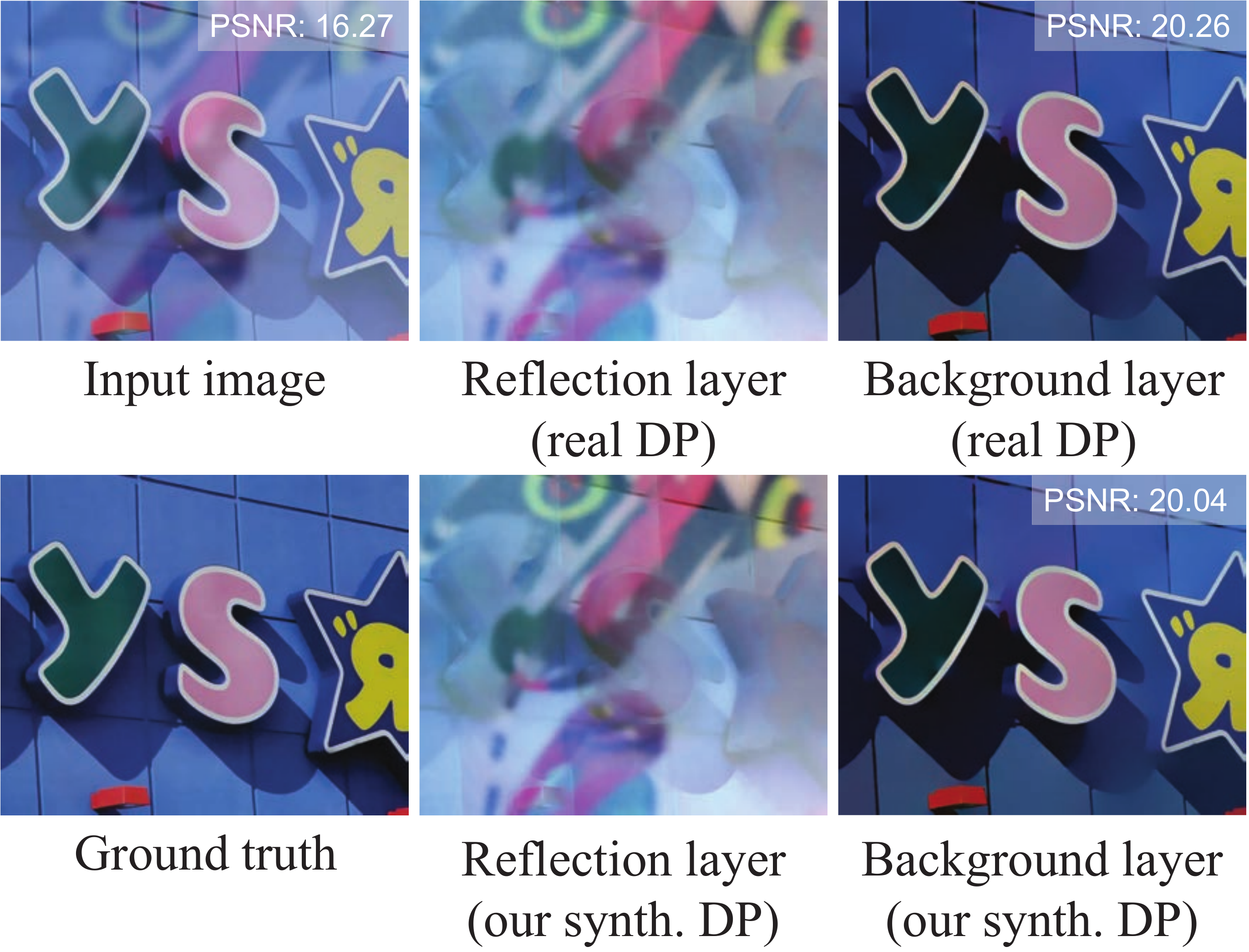}
    \vspace{-2mm}
    \caption{Our synthetic DP views can be used as input for DP-based reflection removal of \cite{punnappurath2019reflection}, and are able to produce results on par with those using real DP data.\label{fig:reflection_removal}}
    \vspace{-2mm}
\end{figure}

\subsection{Ablation study} \label{sec:ablation}

As explained in Sec.\ \ref{sec:method}, our method is a multi-task framework that is suitable not only for reducing defocus blur but also for predicting DP views of the input single image. Our multi-task framework allows our method to improve the results of each task, as they are inherently correlated. Intuitively, we can re-design our framework by training a single model for each task separately. Table \ref{table:ablation_study} shows the results of training a single model (with approximately the same capacity of our multi-task framework) on each task separately. Table \ref{table:ablation_study} also shows the results of training both single and multi-task frameworks with and without our DP-based loss functions introduced in Sec.\ \ref{sec:loss}. As shown, our multi-task framework with our introduced loss functions achieves the best results. Supplementary material provides an additional ablation study on our design of the multi-task architecture.

\subsection{DP-view synthesis and application}

An interesting side effect of our multi-task network is the ability to perform view synthesis.  For instance, we can generate an aesthetically realistic image motion by synthesizing a multi-view version of a given single image. As discussed in Sec.\ \ref{sec:DP_image_formation}, the DP two sub-aperture views of the scene depend on the sensor's orientation and, in this paper, our DLDP dataset contains left/right DP pairs, and consequently our network synthesizes the horizontal DP disparity. We can synthesize additional views with different ``DP disparity'' by rotating the input image before feeding it to our network by a $45\degree$ clockwise step three times (i.e., $45\degree$, $90\degree$, $135\degree$). This allows us to produce a smooth image motion from the reconstructed eight views as shown in Fig.~\ref{fig:multi_view}. We refer to this DP-based view synthesis and image motion as \emph{new image motion attribute} (NIMAT) effect. See supplementary material for additional examples of our novel NIMAT effect. Additionally, we provided another application of our NIMAT effect in~\cite{abuolaim2022multi}. Specifically, we introduced a modification on the blur synthesis procedure in the smartphones' portrait mode to produce a bokeh photo along with our new NIMAT effect.

As discussed in Sec.\ \ref{sec:intro}, DP data has been used for different computer vision tasks.  Here we show using our synthesized DP views can be leveraged for a DP-based method in the absence of real DP data. We validate this idea of using our reconstructed DP views as a proxy to DP data on the reflection removal and defocus deblurring tasks. Specifically, we processed input real DP data and our generated DP data using the DP-based reflection removal (DPRR)~\cite{punnappurath2019reflection}  and defocus deblurring (DPDNet)~\cite{abuolaim2020defocus} methods. As shown in Fig.\ \ref{fig:reflection_removal}, utilizing our synthetic DP views produces approximately the same high-quality result as using DPRR\cite{punnappurath2019reflection} on real DP data. This allows us to achieve better reflection removal results, while still requiring only \textit{a single input image}, compared to other methods for reflection removal, as shown in Table \ref{table:reflection_removal}. As for DPDNet, Table~\ref{table:dp_deblurring} shows that DPDNet tested with our generated DP views has on-par results compared to the one tested with real DP views.

\section{Conclusion}
We have demonstrated that a DNN trained for the purpose of single-image defocus deblurring can be improved by incorporating the additional task of synthesizing the two DP views associated with the input image.  One benefit of this approach is that capturing data for the DP view synthesis task is easy to perform and requires no special capture setup.  This is contrasted with a conventional approach that requires careful capture of sharp/blurred image pairs for the deblurring task.  This multi-task strategy is able to improve deblurring results by close to 1dB in terms of PSNR. As an added benefit, we introduced the novel NIMAT effect and showed that our DNN is able to perform realistic view synthesis that can be used for tasks such as reflection removal. Our DLDP dataset, code, and trained models are publicly available at~\url{https://github.com/Abdullah-Abuolaim/multi-task-defocus-deblurring-dual-pixel-nimat}.

\paragraph{Acknowledgments}
A big thanks to David Ampofo for his amazing photography skills in collecting the images for our new DLDP dataset.
\clearpage

\setcounter{section}{0}
\setcounter{footnote}{0}
\setcounter{figure}{0}
\setcounter{table}{0}
\setcounter{equation}{0}

\newcommand{\hbAppendixPrefix}{A}
\renewcommand{\thesection}{S\arabic{section}}
\renewcommand{\thetable}{S\arabic{table}}
\renewcommand{\thefigure}{S\arabic{figure}}

\twocolumn[\centerline{\Large{\textbf{Supplemental Material}}}]

\vspace*{0.7cm}

\begin{figure*}
     \begin{subfigure}[]{\textwidth}
         \includegraphics[width=\textwidth]{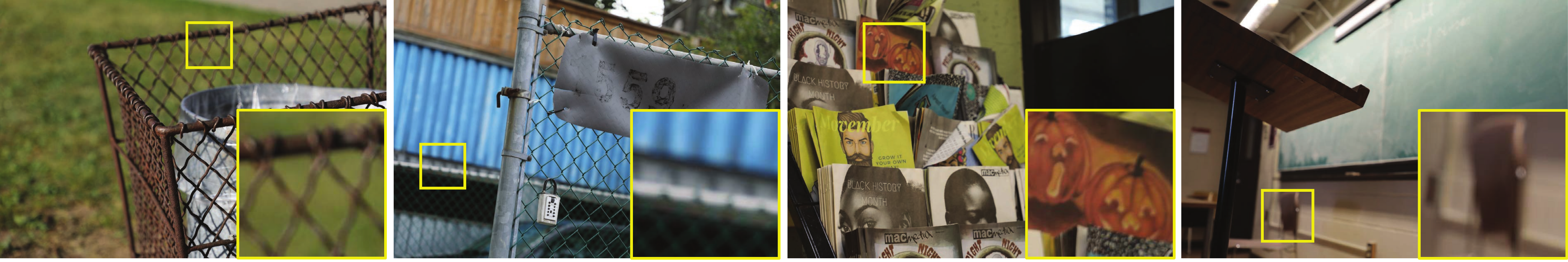}
         \vspace{-6.3mm}
         \caption{Input}
     \end{subfigure}
     
     \vspace{1mm}
     \begin{subfigure}[]{\textwidth}
         \includegraphics[width=\textwidth]{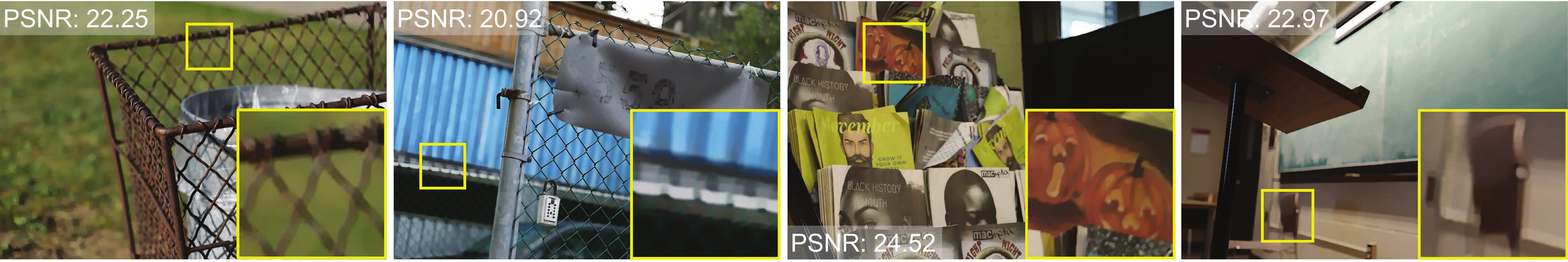}
         \vspace{-6.3mm}
         \caption{EBDB \cite{karaali2017edge}}
     \end{subfigure}
     
     \vspace{1mm}
     \begin{subfigure}[]{\textwidth}
         \includegraphics[width=\textwidth]{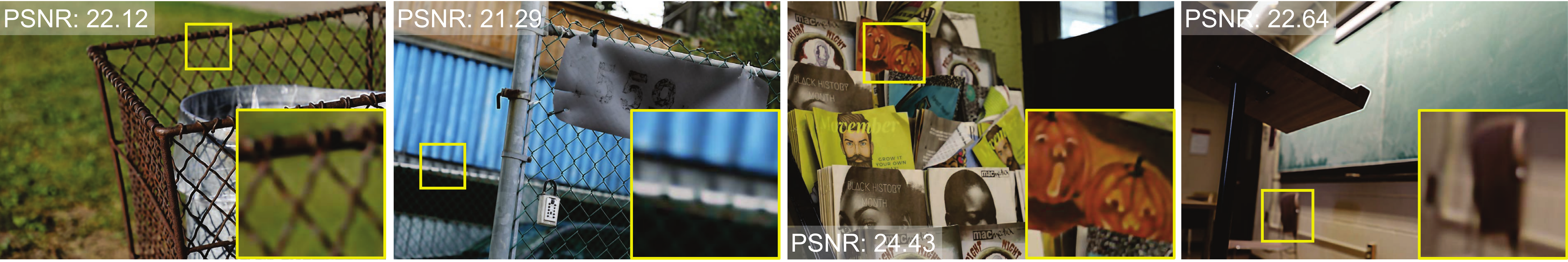}
         \vspace{-6.3mm}
         \caption{DMENet \cite{lee2019deep}}
     \end{subfigure}
     
    \vspace{1mm}
     \begin{subfigure}[]{\textwidth}
         \includegraphics[width=\textwidth]{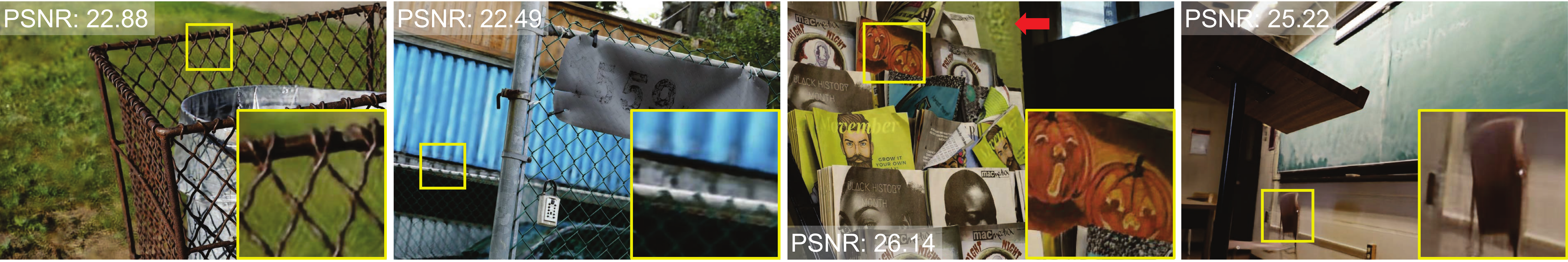}
         \vspace{-6.3mm}
         \caption{DPDNet (single) \cite{abuolaim2020defocus}}
     \end{subfigure}
     
     \vspace{1mm}
     \begin{subfigure}[]{\textwidth}
         \includegraphics[width=\textwidth]{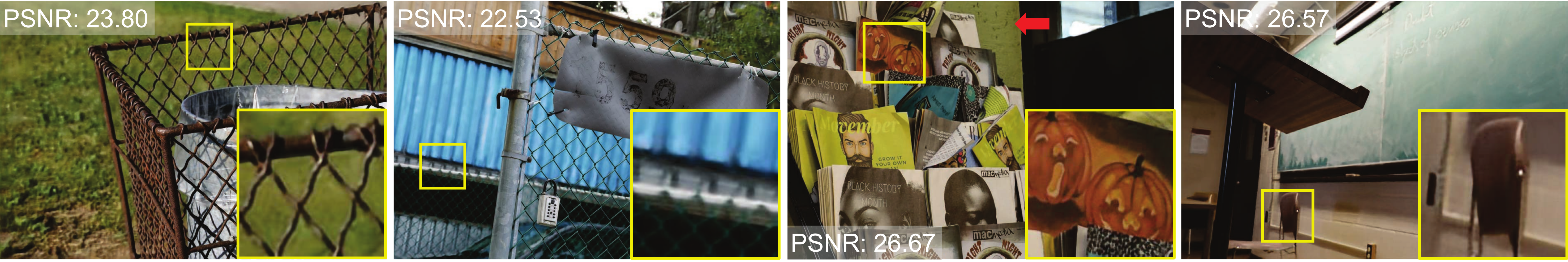}
         \vspace{-6.3mm}
         \caption{Ours}
     \end{subfigure}
     
     \vspace{1mm}
     \begin{subfigure}[]{\textwidth}
         \includegraphics[width=\textwidth]{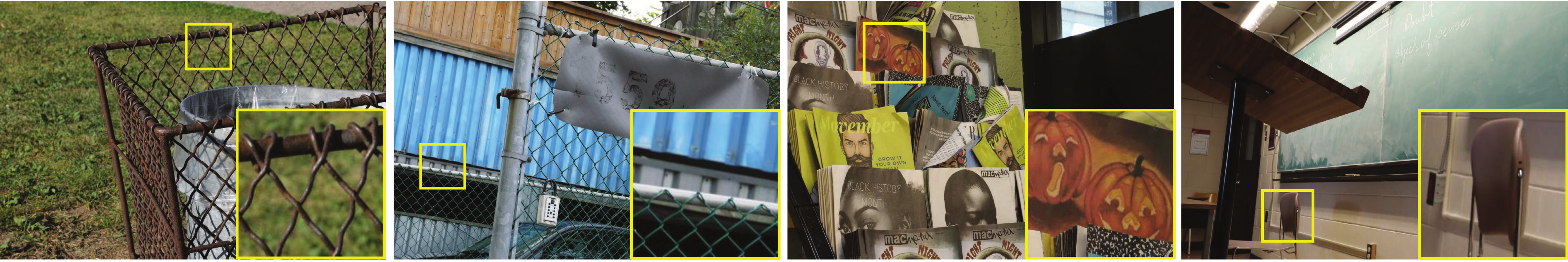}
         \vspace{-6.3mm}
         \caption{Ground truth}
     \end{subfigure}
    
    \caption{Additional qualitative comparisons with other single-image defocus deblurring methods on the test set of the Canon DP dataset \cite{abuolaim2020defocus}. We compare our results with the following single-image defocus deblurring methods:  EBDB \cite{karaali2017edge}, DMENet \cite{lee2019deep}, and DPDNet (single) \cite{abuolaim2020defocus}. Note that DPDNet was originally introduced to use DP images as input, but the authors in~\cite{abuolaim2020defocus} also provided the same model trained on a single image, denoted as DPDNet (single). Our method produces the best quantitative and arguably best qualitative results.}
        \label{fig:qualitative_supp}
\end{figure*}

\begin{figure*}
    \centering
     \begin{subfigure}[]{0.245\textwidth}
         \includegraphics[width=\textwidth]{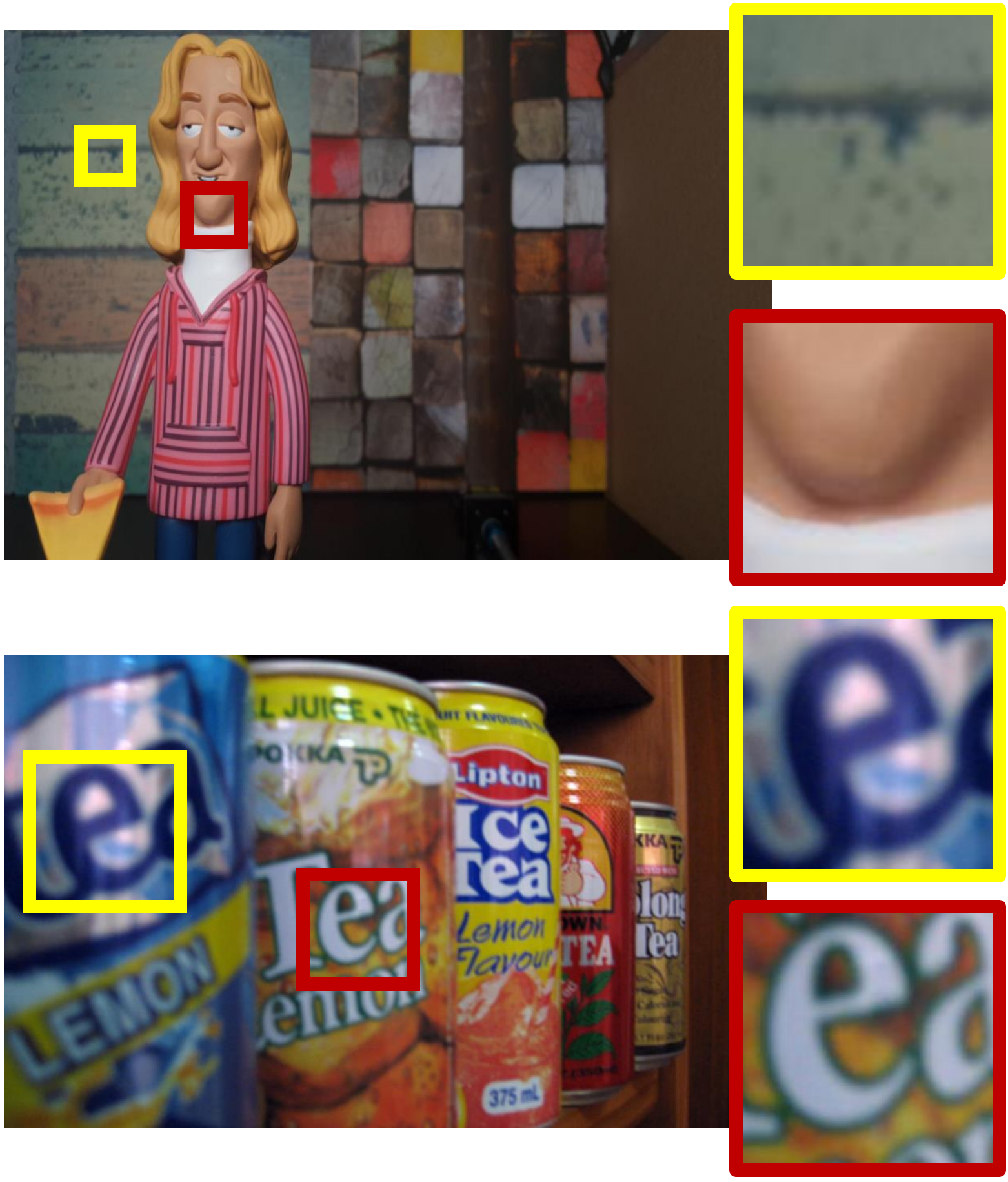}

         \caption{Input}
     \end{subfigure}
     \begin{subfigure}[]{0.245\textwidth}
         \includegraphics[width=\textwidth]{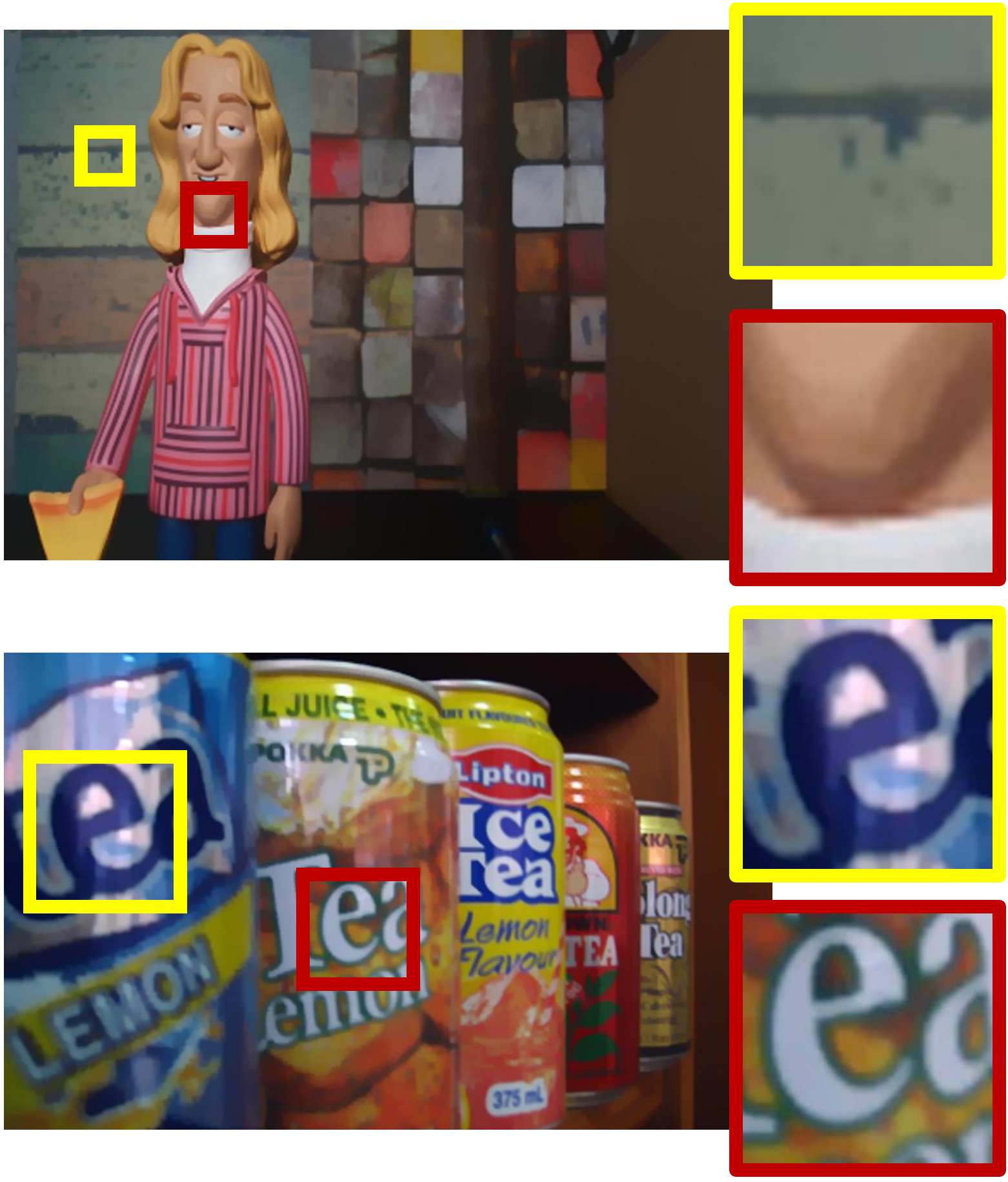}

         \caption{EBDB \cite{karaali2017edge}}
     \end{subfigure}
     \begin{subfigure}[]{0.245\textwidth}
         \includegraphics[width=\textwidth]{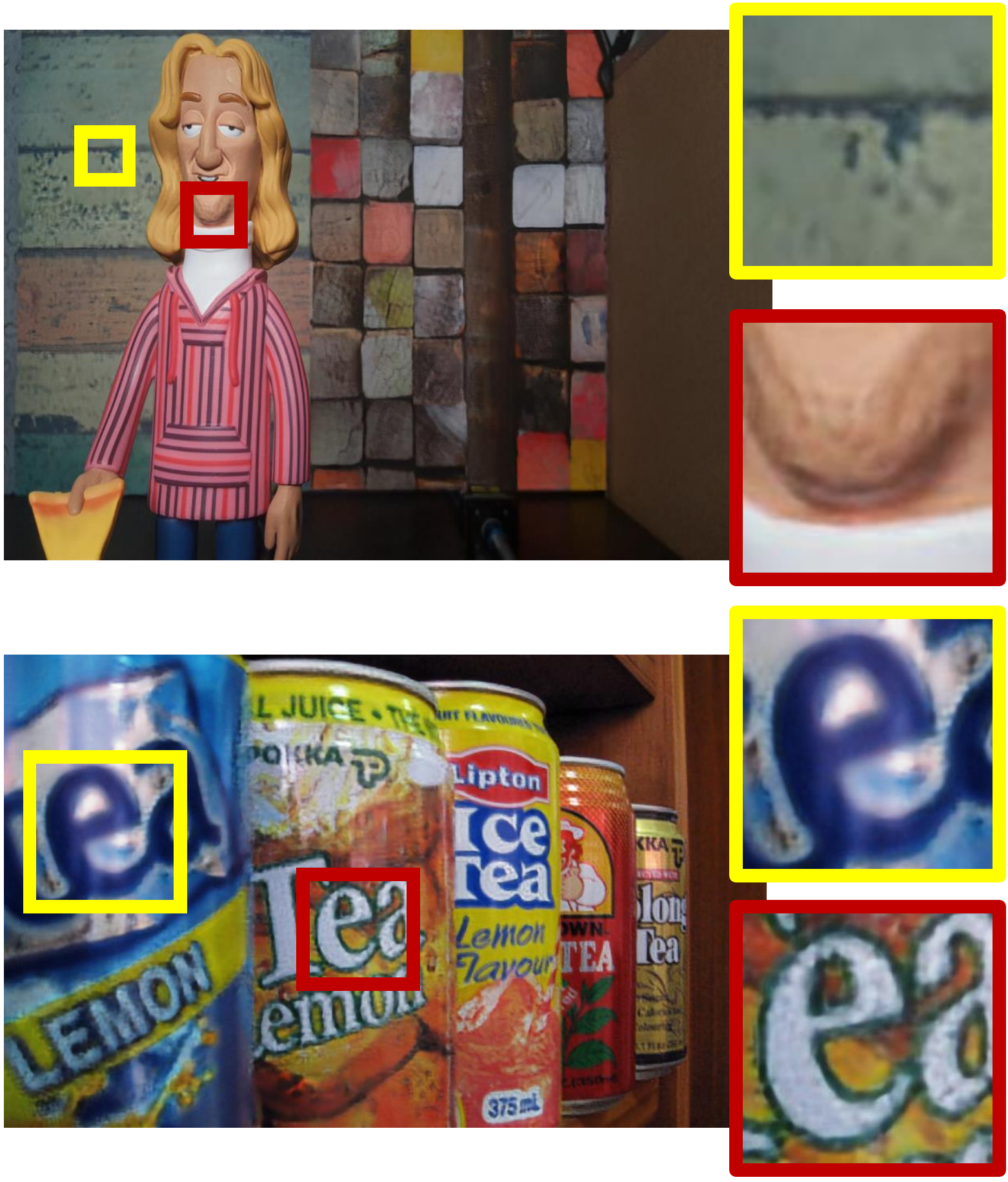}

         \caption{DPDNet (single) \cite{abuolaim2020defocus}}
     \end{subfigure}
    \vspace{1mm}
     \begin{subfigure}[]{0.245\textwidth}
         \includegraphics[width=\textwidth]{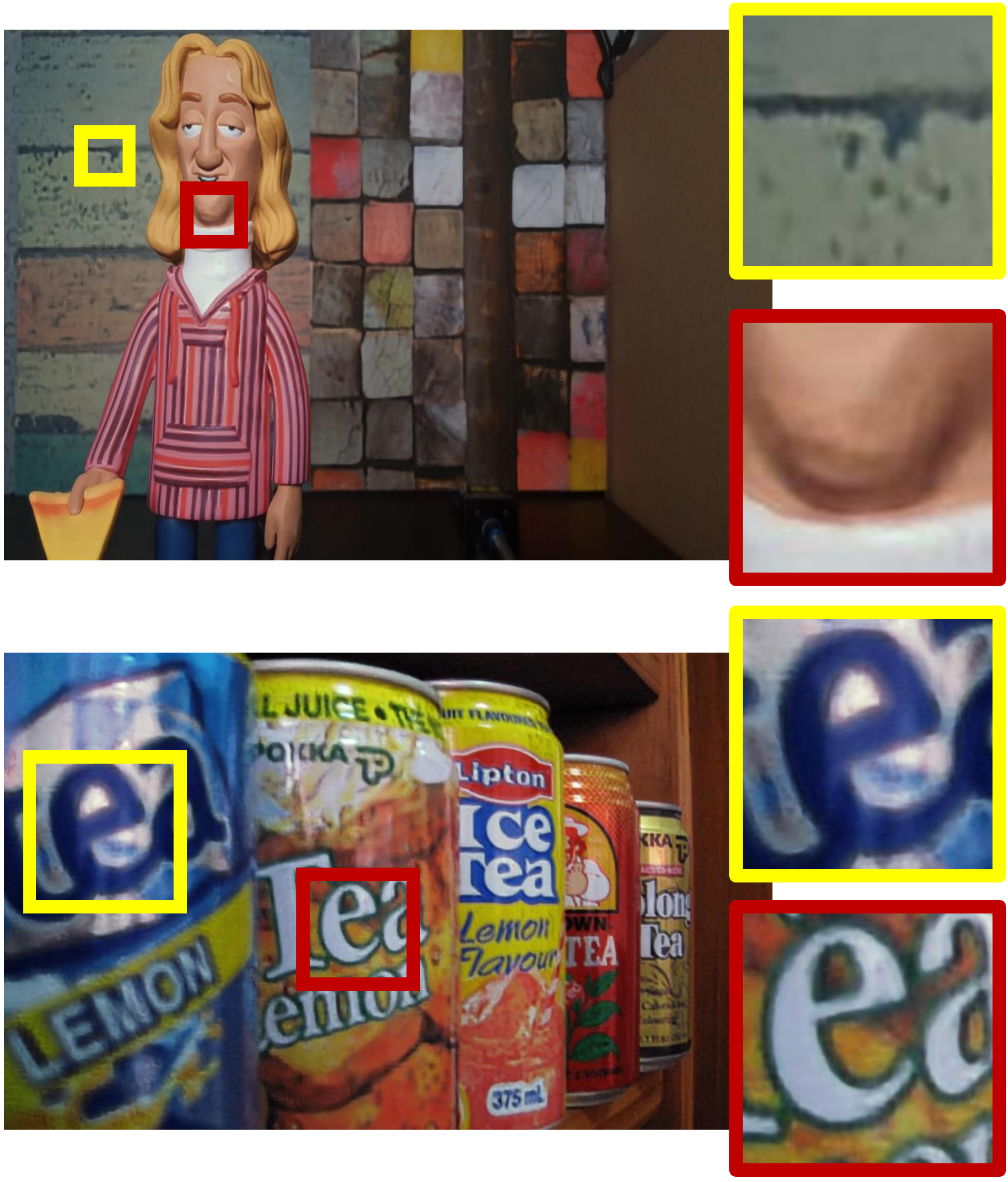}

         \caption{Ours}
     \end{subfigure}

    \caption{Qualitative comparison with single-image defocus deblurring methods on other camera devices (top: Samsung S6, bottom: Flickr image [erasmusa CC BY-NC 2]. We compare our results with the following single-image defocus deblurring methods:  EBDB \cite{karaali2017edge} and DPDNet (single) \cite{abuolaim2020defocus}. Note that DPDNet was originally introduced to use DP images as input, but the authors in~\cite{abuolaim2020defocus} also provided the same model trained on a single image, denoted as DPDNet (single). Our method generalizes well for unseen cameras during the training stage and produces arguably best qualitative results compared with other methods.}
        \label{fig:qualitative_supp_other}
\end{figure*}

\begin{figure*}
    \centering
     \begin{subfigure}[]{0.49\textwidth}
         \includegraphics[width=\textwidth]{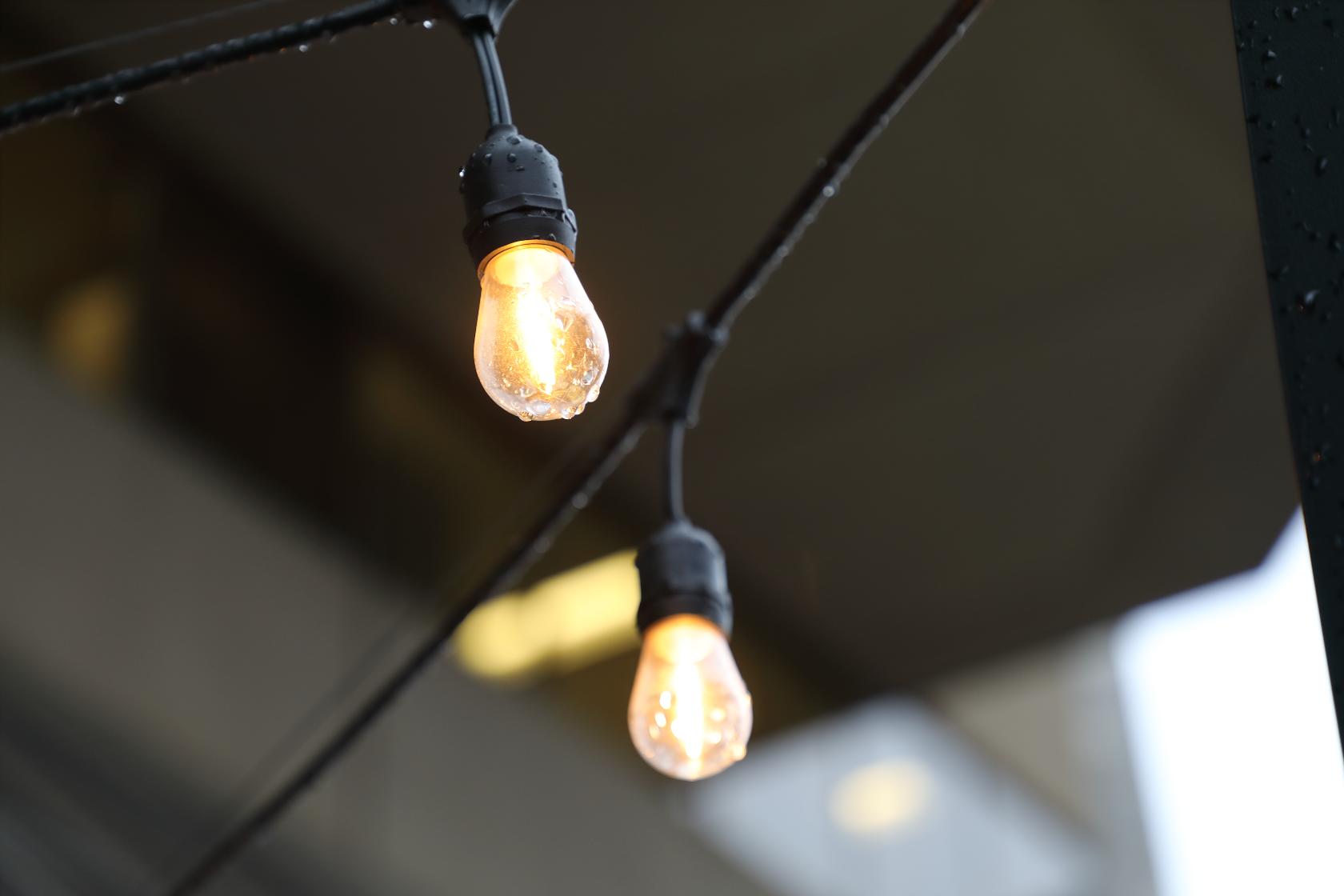}

         \caption{Input}
     \end{subfigure}
     \begin{subfigure}[]{0.49\textwidth}
         \animategraphics[width=\textwidth]{12}{supp_dataset_motion_1/001_}{0}{55}

         \caption{Our generated NIMAT effect}
     \end{subfigure}
     
     \begin{subfigure}[]{0.49\textwidth}
         \animategraphics[width=\textwidth]{3}{supp_dataset_p_1/1_p_}{1}{12}

         \caption{Our DP views}
     \end{subfigure}
     \begin{subfigure}[]{0.49\textwidth}
         \animategraphics[width=\textwidth]{3}{supp_dataset_gt_1/1_g_}{1}{12}

         \caption{GT DP views}
     \end{subfigure}
    \caption{An example from our DLDP dataset. (a) Input combined image $\mathbf{I}_c$. (b) Our novel NIMAT effect generated using the proposed MDP. (c) Animated results of our synthesized DP views. (d) Animated ground truth DP views. Our MDP is able to generate high-quality eight/DP views. {\bf Note: (b), (c), and (d) are animated figures; click on the figure to start the animation. It is recommended to open this PDF in Adobe Acrobat Reader to work properly.}}
        \label{fig:multi-view_dp_dataset_1}
\end{figure*}

\begin{figure*}
    \centering
     \begin{subfigure}[]{0.49\textwidth}
         \includegraphics[width=\textwidth]{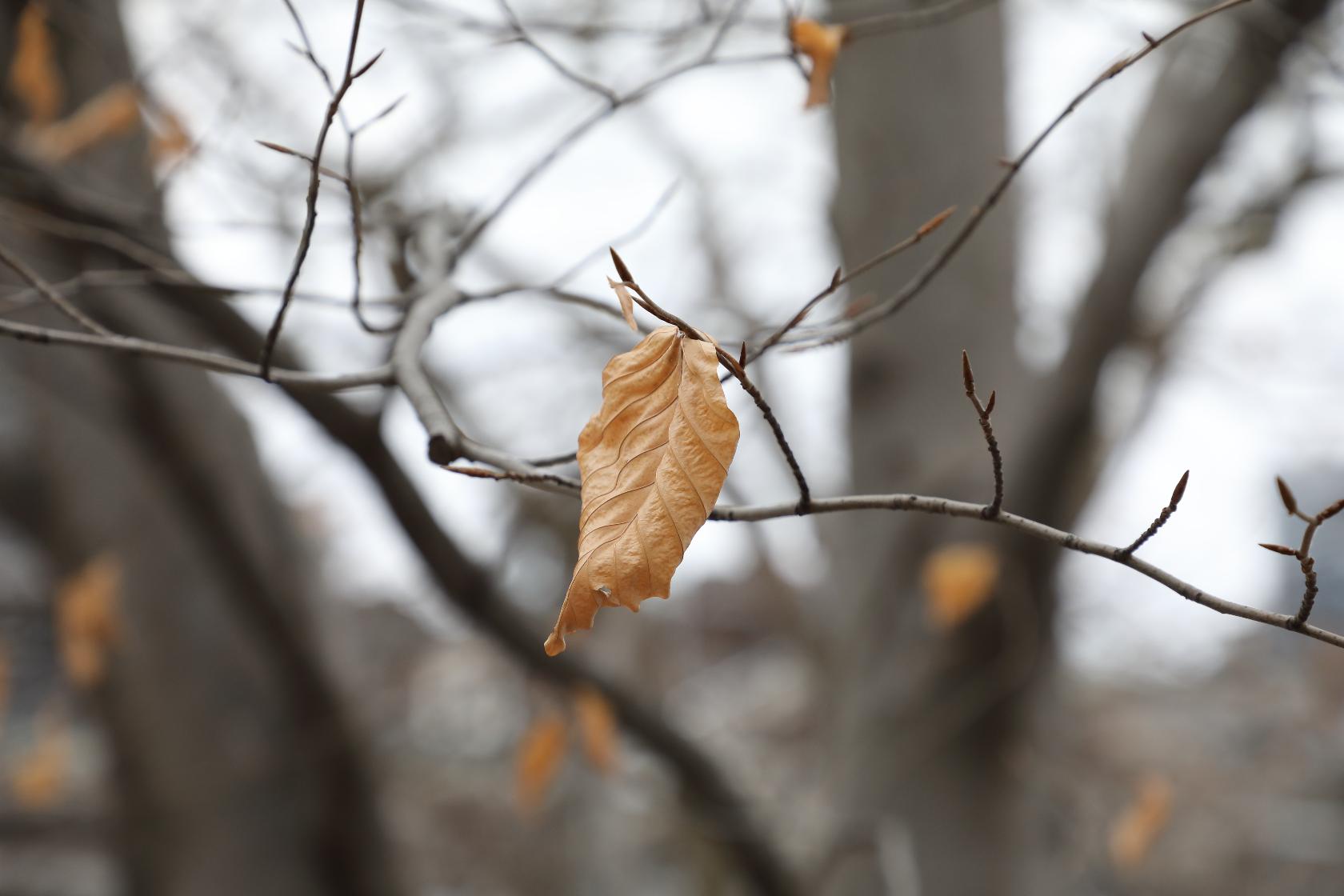}

         \caption{Input}
     \end{subfigure}
     \begin{subfigure}[]{0.49\textwidth}
         \animategraphics[width=\textwidth]{12}{supp_dataset_motion_3/003_}{0}{55}

         \caption{Our generated NIMAT effect}
     \end{subfigure}
     
     \begin{subfigure}[]{0.49\textwidth}
         \animategraphics[width=\textwidth]{3}{supp_dataset_p_3/3_p_}{1}{12}

         \caption{Our DP views}
     \end{subfigure}
     \begin{subfigure}[]{0.49\textwidth}
         \animategraphics[width=\textwidth]{3}{supp_dataset_gt_3/3_g_}{1}{12}

         \caption{GT DP views}
     \end{subfigure}
    \caption{Additional example from our DLDP dataset. (a) Input combined image $\mathbf{I}_c$. (b) Our novel NIMAT effect generated using the proposed MDP. (c) Animated results of our synthesized DP views. (d) Animated ground truth DP views. Our MDP is able to generate high-quality eight/DP views. {\bf Note: (b), (c), and (d) are animated figures; click on the figure to start the animation. It is recommended to open this PDF in Adobe Acrobat Reader to work properly.}\label{fig:multi-view_dp_dataset_2}}
\end{figure*}

\begin{figure*}
    \centering
     \begin{subfigure}[]{0.49\textwidth}
         \includegraphics[width=\textwidth]{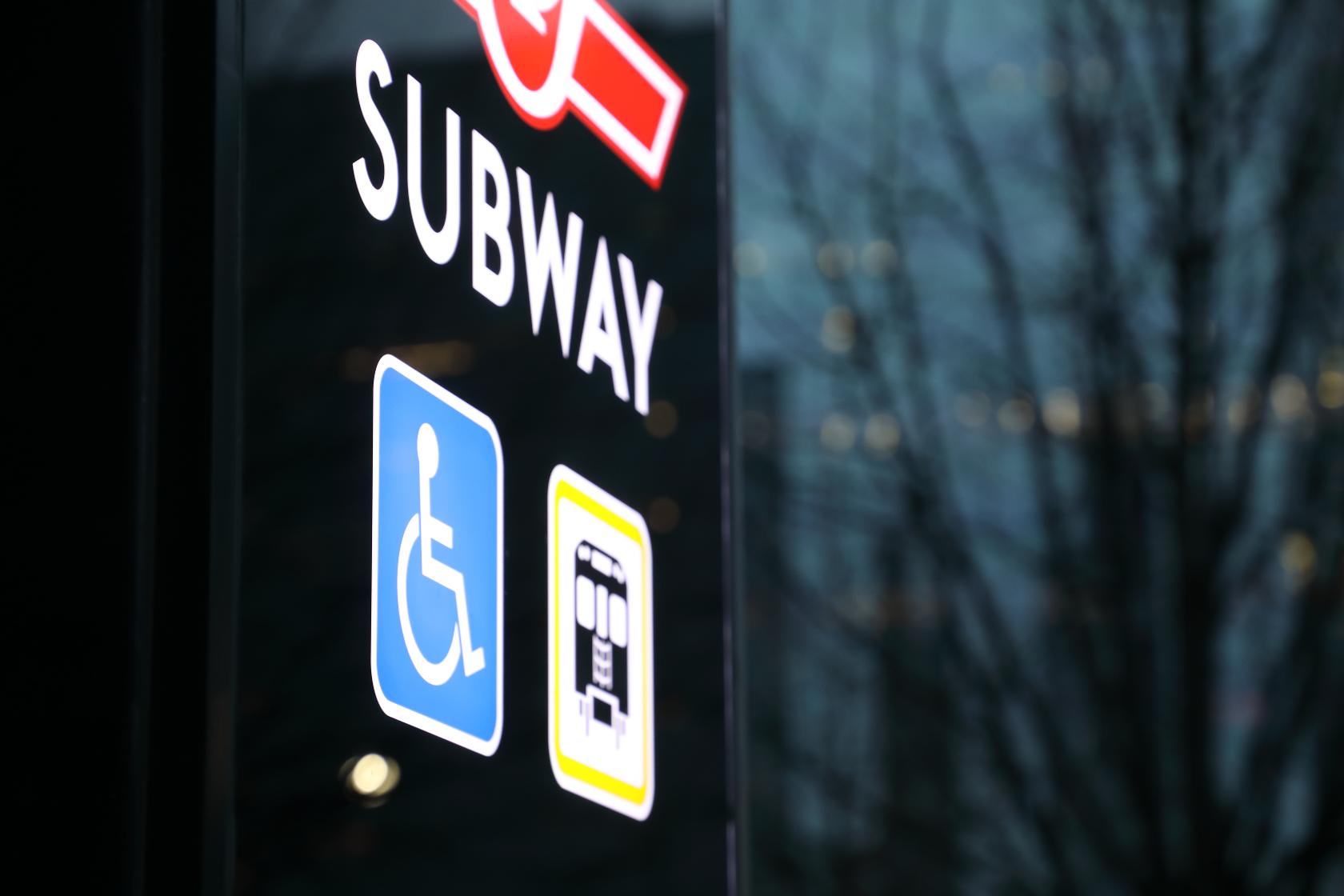}

         \caption{Input}
     \end{subfigure}
     \begin{subfigure}[]{0.49\textwidth}
         \animategraphics[width=\textwidth]{12}{supp_dataset_motion_5/005_}{0}{55}

         \caption{Our generated NIMAT effect}
     \end{subfigure}
     
     \begin{subfigure}[]{0.49\textwidth}
         \animategraphics[width=\textwidth]{3}{supp_dataset_p_5/5_p_}{1}{12}

         \caption{Our DP views}
     \end{subfigure}
     \begin{subfigure}[]{0.49\textwidth}
         \animategraphics[width=\textwidth]{3}{supp_dataset_gt_5/5_g_}{1}{12}

         \caption{GT DP views}
     \end{subfigure}
    \caption{Additional example from our DLDP dataset. (a) Input combined image $\mathbf{I}_c$. (b) Our novel NIMAT effect generated using the proposed MDP. (c) Animated results of our synthesized DP views. (d) Animated ground truth DP views. Our MDP is able to generate high-quality eight/DP views. {\bf Note: (b), (c), and (d) are animated figures; click on the figure to start the animation. It is recommended to open this PDF in Adobe Acrobat Reader to work properly.}\label{fig:multi-view_dp_dataset_3}}
\end{figure*}

\begin{figure*}
    \centering
     \begin{subfigure}[]{0.49\textwidth}
         \includegraphics[width=\textwidth]{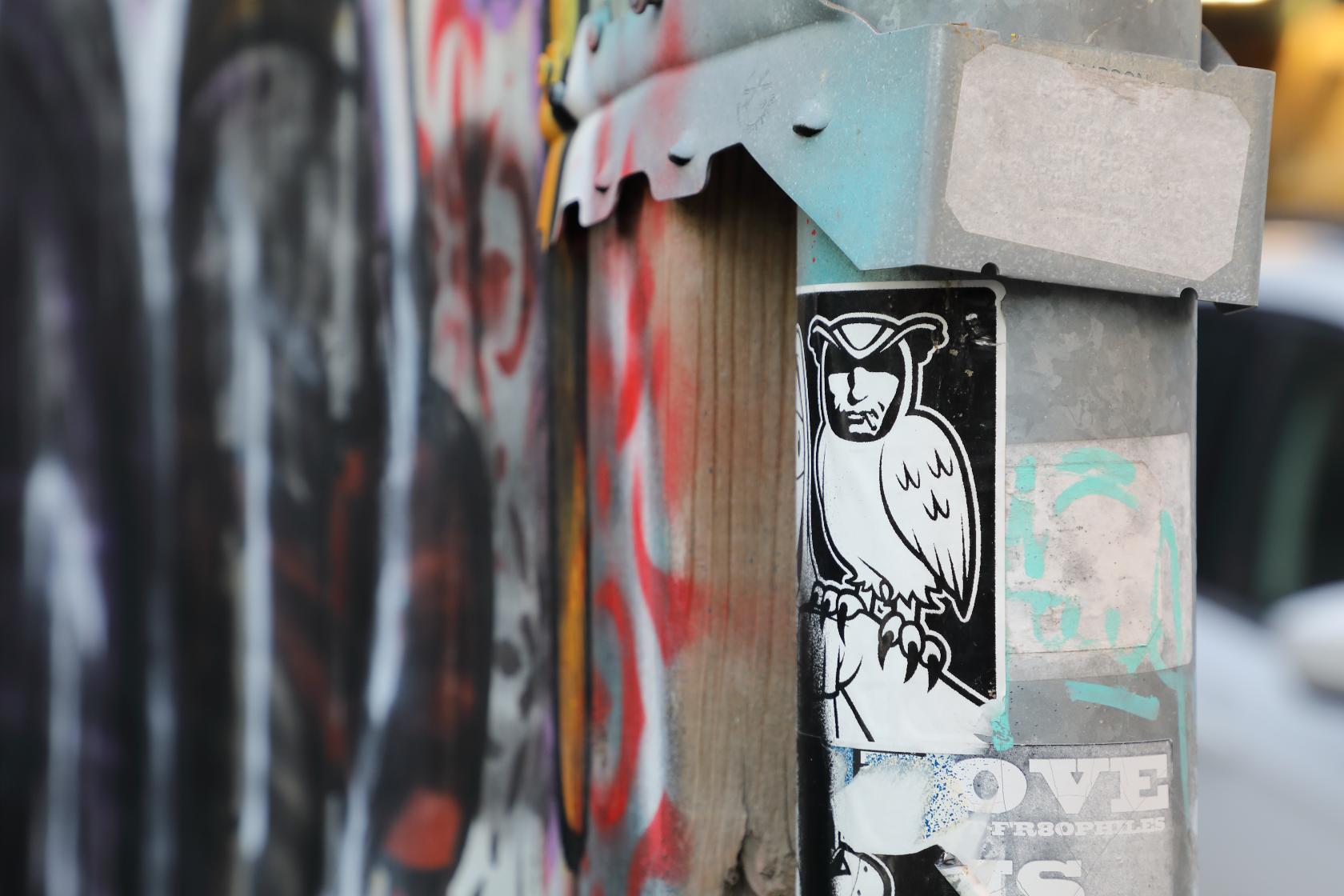}

         \caption{Input}
     \end{subfigure}
     \begin{subfigure}[]{0.49\textwidth}
         \animategraphics[width=\textwidth]{12}{supp_dataset_motion_7/007_}{0}{55}

         \caption{Our generated NIMAT effect}
     \end{subfigure}
     
     \begin{subfigure}[]{0.49\textwidth}
         \animategraphics[width=\textwidth]{3}{supp_dataset_p_7/7_p_}{1}{12}

         \caption{Our DP views}
     \end{subfigure}
     \begin{subfigure}[]{0.49\textwidth}
         \animategraphics[width=\textwidth]{3}{supp_dataset_gt_7/7_g_}{1}{12}

         \caption{GT DP views}
     \end{subfigure}
    \caption{Additional example from our DLDP dataset. (a) Input combined image $\mathbf{I}_c$. (b) Our novel NIMAT effect generated using the proposed MDP. (c) Animated results of our synthesized DP views. (d) Animated ground truth DP views. Our MDP is able to generate high-quality eight/DP views. {\bf Note: (b), (c), and (d) are animated figures; click on the figure to start the animation. It is recommended to open this PDF in Adobe Acrobat Reader to work properly.}}
        \label{fig:multi-view_dp_dataset_4}
\end{figure*}

This supplementary material provides an ablation study (Sec.~\ref{sec:supp-ablation}) to show the effectiveness of the multi-decoder stitching at the middle stage vs. other options (i.e., late-stage and no stitching) of the proposed multi-task dual-pixel network (MDP). Additional qualitative results are also provided as follows:
\begin{itemize}
    \item Fig.~\ref{fig:qualitative_supp} provides a qualitative comparison with other single-image defocus deblurring methods tested on the Canon DP dataset~\cite{abuolaim2020defocus}. These methods are: the just noticeable defocus blur method (JNB) \cite{shi2015just}, the edge-based defocus blur estimation method (EBDB) \cite{karaali2017edge}, the deep defocus map estimation method (DMENet) \cite{lee2019deep}, and the DPDNet (single) \cite{abuolaim2020defocus}.
    
    \item Fig.~\ref{fig:qualitative_supp_other} demonstrates qualitatively the deblurring generalization ability of our proposed MDP.  In this experiment, MDP is trained only on the Canon data, but tested on images from other cameras.
    
    \item Figs.~\ref{fig:multi-view_dp_dataset_1}, \ref{fig:multi-view_dp_dataset_2}, \ref{fig:multi-view_dp_dataset_3}, and \ref{fig:multi-view_dp_dataset_4} show examples from our newly captured DLDP dataset along with results of high-quality reconstructed DP views and image motion of the synthesized eight views. The eight views are generated based on the description in Sec.~{\color{red}4.4} of the main paper, and visualized as image motion by alternating through eight views (i.e., NIMAT effect).
    
    \item Fig.~\ref{fig:image_motion} shows qualitatively how our proposed MDP is able to generalize for other cameras. In this experiment, we synthesize eight views from a single-input image captured by other cameras. The eight views are generated based on the description in Sec.~{\color{red}4.4} of the main paper, and visualized as image motion by alternating the eight views (i.e., NIMAT effect).
\end{itemize}

\section{Ablation study of multi-decoder stitching \label{sec:supp-ablation}}
In this section, we investigate the utility of having multiple weight sharing stages by introducing a variation of MDP network with different multi-decoder stitching options: (1) {\it no stitching} that makes the latent space $\mathcal{X}$ the only weight sharing stage, (2) {\it late-stage stitching} at the last block, and (3) the original proposed MDP with {\it middle-stage stitching}. We report the results of MDP variations in Table~\ref{table:ablation_study_stitching}. The training procedure followed is the same as for all MDP variations as described in Sec.~{\color{red}4.1} of the main paper.

The results in Table~\ref{table:ablation_study_stitching} show that {\it middle-stage stitching} achieves the best results as it allows weight sharing at multiple stages compared with the {\it no stitching} variation. On the other hand, there is a noticeable drop in the deblurring performance when {\it late-stage stitching} is applied as the sharpness of the deblurring decoder (i.e., $\mathbf{Dec}_s$) is affected by the half-PSF blur present in feature maps of the synthesized DP views (i.e., $\mathbf{Dec}_l$ and $\mathbf{Dec}_r$) at this later stage.

\begin{table}[t!]
\vspace*{1.22cm}
\caption{This table reports results on an ablation study performed to examine the effectiveness of the multi-decoder stitching design for defocus deblurring. Three variations of our proposed MDP based on three different stitching options: (1) no stitching, (2) late-stage stitching and (3) middle-stage stitching. Results reported are on the Canon DP deblurring dataset \cite{abuolaim2020defocus}.\label{table:ablation_study_stitching}}
\vspace{-2mm}
\centering
\scalebox{0.84}{
\begin{tabular}{l|ccc}
MDP variation & PSNR $\uparrow$ & SSIM $\uparrow$ & MAE $\downarrow$ \\ \hline
MDP (no stitching) & 25.03 & 0.757 & 0.042 \\
MDP (late-stage stitching) & 25.16 & 0.759 & 0.041 \\\hdashline
MDP (middle-stage stitching) & {\bf 25.35} & {\bf 0.763} & {\bf 0.040} \\
\end{tabular}}
\vspace{-2mm}
\end{table}

\begin{figure*}
    \centering
     \begin{subfigure}[]{0.45\textwidth}
         \animategraphics[width=\textwidth]{12}{animation_example_supp_1/image_motion_supp_001_}{0}{55}

     \end{subfigure}\hspace{2mm}
     \begin{subfigure}[]{0.45\textwidth}
         \animategraphics[width=\textwidth]{12}{animation_example_supp_3/image_motion_supp_003_}{0}{55}

     \end{subfigure}

     \vspace{4mm}

    \begin{subfigure}[]{0.45\textwidth}
        \animategraphics[width=\textwidth]{12}{animation_example_supp_7/image_motion_supp_007_}{0}{55}

     \end{subfigure}\hspace{2mm}
    \begin{subfigure}[]{0.45\textwidth}
        \animategraphics[width=\textwidth]{12}{animation_example_supp_6/image_motion_supp_006_}{0}{55}

     \end{subfigure}

    \caption{Our novel NIMAT effect. Multi-view synthesis results of our proposed MDP applied to other cameras than the Canon 5D DSLR camera (used for training). These results are synthesized from a single input image captured by new camera devices, in which they do not have the ground truth DP views. Our MDP produces high-quality eight views that can be used to create an aesthetically pleasing NIMAT effect. Furthermore, these results demonstrate a good generalization ability of our MDP as it can provide high-quality views from images that are captured by unseen camera device during the training stage. {\bf Note: these images are animated; click on the image to start the animation. It is recommended to open this PDF in Adobe Acrobat Reader to work properly.}}
\label{fig:image_motion}
\end{figure*}

\clearpage
\clearpage

{\small
\bibliographystyle{ieee_fullname}
\bibliography{arXiv_2022_multi_task_dp_defocus_deblurring}

\begin{thebibliography}{10}\itemsep=-1pt

\bibitem{abuolaim2022multi}
Abdullah Abuolaim, Mahmoud Afifi, and Michael~S Brown.
\newblock Multi-view motion synthesis via applying rotated dual-pixel blur
  kernels.
\newblock In {\em WACV Workshops}, 2022.

\bibitem{abuolaim2020defocus}
Abdullah Abuolaim and Michael~S Brown.
\newblock Defocus deblurring using dual-pixel data.
\newblock In {\em ECCV}, 2020.

\bibitem{abuolaim2020online}
Abdullah Abuolaim and Michael~S Brown.
\newblock Online lens motion smoothing for video autofocus.
\newblock In {\em WACV}, 2020.

\bibitem{abuolaim2018revisiting}
Abdullah Abuolaim, Abhijith Punnappurath, and Michael~S Brown.
\newblock Revisiting autofocus for smartphone cameras.
\newblock In {\em ECCV}, 2018.

\bibitem{abuolaim2021ntire}
Abdullah Abuolaim, Radu Timofte, Michael~S Brown, et~al.
\newblock {NTIRE 2021} challenge for defocus deblurring using dual-pixel
  images: Methods and results.
\newblock In {\em CVPR Workshops}, 2021.

\bibitem{d2016non}
Laurent D’Andr{\`e}s, Jordi Salvador, Axel Kochale, and Sabine S{\"u}sstrunk.
\newblock Non-parametric blur map regression for depth of field extension.
\newblock {\em TIP}, 25(4):1660--1673, 2016.

\bibitem{fish1995blind}
DA Fish, AM Brinicombe, ER Pike, and JG Walker.
\newblock Blind deconvolution by means of the richardson--lucy algorithm.
\newblock {\em Journal of the Optical Society of America (A)}, 12(1):58--65,
  1995.

\bibitem{garg2019learning}
Rahul Garg, Neal Wadhwa, Sameer Ansari, and Jonathan~T Barron.
\newblock Learning single camera depth estimation using dual-pixels.
\newblock In {\em ICCV}, 2019.

\bibitem{girshick2015fast}
Ross Girshick.
\newblock Fast {R}-{CNN}.
\newblock In {\em CVPR}, 2015.

\bibitem{guo2014robust}
Xiaojie Guo, Xiaochun Cao, and Yi Ma.
\newblock Robust separation of reflection from multiple images.
\newblock In {\em CVPR}, 2014.

\bibitem{he2015delving}
Kaiming He, Xiangyu Zhang, Shaoqing Ren, and Jian Sun.
\newblock Delving deep into rectifiers: Surpassing human-level performance on
  imagenet classification.
\newblock In {\em ICCV}, 2015.

\bibitem{karaali2017edge}
Ali Karaali and Claudio~Rosito Jung.
\newblock Edge-based defocus blur estimation with adaptive scale selection.
\newblock {\em TIP}, 27(3):1126--1137, 2018.

\bibitem{kingma2014adam}
Diederik~P Kingma and Jimmy Ba.
\newblock Adam: A method for stochastic optimization.
\newblock {\em arXiv preprint arXiv:1412.6980}, 2014.

\bibitem{krishnan2009fast}
Dilip Krishnan and Rob Fergus.
\newblock Fast image deconvolution using hyper-laplacian priors.
\newblock In {\em NeurIPS}, 2009.

\bibitem{lee2019deep}
Junyong Lee, Sungkil Lee, Sunghyun Cho, and Seungyong Lee.
\newblock Deep defocus map estimation using domain adaptation.
\newblock In {\em CVPR}, 2019.

\bibitem{lee2021iterative}
Junyong Lee, Hyeongseok Son, Jaesung Rim, Sunghyun Cho, and Seungyong Lee.
\newblock Iterative filter adaptive network for single image defocus
  deblurring.
\newblock In {\em CVPR}, 2021.

\bibitem{levin2011understanding}
Anat Levin, Yair Weiss, Fredo Durand, and William~T Freeman.
\newblock Understanding blind deconvolution algorithms.
\newblock {\em TPAMI}, 33(12):2354--2367, 2011.

\bibitem{li2013exploiting}
Yu Li and Michael~S Brown.
\newblock Exploiting reflection change for automatic reflection removal.
\newblock In {\em ICCV}, 2013.

\bibitem{liu2019end}
Shikun Liu, Edward Johns, and Andrew~J Davison.
\newblock End-to-end multi-task learning with attention.
\newblock In {\em CVPR}, 2019.

\bibitem{mao2016image}
Xiaojiao Mao, Chunhua Shen, and Yu-Bin Yang.
\newblock Image restoration using very deep convolutional encoder-decoder
  networks with symmetric skip connections.
\newblock In {\em NeurIPS}, 2016.

\bibitem{nah2019recurrent}
Seungjun Nah, Sanghyun Son, and Kyoung~Mu Lee.
\newblock Recurrent neural networks with intra-frame iterations for video
  deblurring.
\newblock In {\em CVPR}, 2019.

\bibitem{pan2020cascaded}
Jinshan Pan, Haoran Bai, and Jinhui Tang.
\newblock Cascaded deep video deblurring using temporal sharpness prior.
\newblock In {\em CVPR}, 2020.

\bibitem{pan2021dual}
Liyuan Pan, Shah Chowdhury, Richard Hartley, Miaomiao Liu, Hongguang Zhang, and
  Hongdong Li.
\newblock Dual pixel exploration: Simultaneous depth estimation and image
  restoration.
\newblock In {\em CVPR}, 2021.

\bibitem{park2017unified}
Jinsun Park, Yu-Wing Tai, Donghyeon Cho, and In So~Kweon.
\newblock A unified approach of multi-scale deep and hand-crafted features for
  defocus estimation.
\newblock In {\em CVPR}, 2017.

\bibitem{punnappurath2020modeling}
Abhijith Punnappurath, Abdullah Abuolaim, Mahmoud Afifi, and Michael~S Brown.
\newblock Modeling defocus-disparity in dual-pixel sensors.
\newblock In {\em ICCP}, 2020.

\bibitem{punnappurath2019reflection}
Abhijith Punnappurath and Michael~S Brown.
\newblock Reflection removal using a dual-pixel sensor.
\newblock In {\em CVPR}, 2019.

\bibitem{ronneberger2015u}
Olaf Ronneberger, Philipp Fischer, and Thomas Brox.
\newblock U-net: Convolutional networks for biomedical image segmentation.
\newblock In {\em MICCAI}, 2015.

\bibitem{shi2015just}
Jianping Shi, Li Xu, and Jiaya Jia.
\newblock Just noticeable defocus blur detection and estimation.
\newblock In {\em CVPR}, 2015.

\bibitem{su2017deep}
Shuochen Su, Mauricio Delbracio, Jue Wang, Guillermo Sapiro, Wolfgang Heidrich,
  and Oliver Wang.
\newblock Deep video deblurring for hand-held cameras.
\newblock In {\em CVPR}, 2017.

\bibitem{tang2012utilizing}
Huixuan Tang and Kiriakos~N Kutulakos.
\newblock Utilizing optical aberrations for extended-depth-of-field panoramas.
\newblock In {\em ACCV}, 2012.

\bibitem{tao2018scale}
Xin Tao, Hongyun Gao, Xiaoyong Shen, Jue Wang, and Jiaya Jia.
\newblock Scale-recurrent network for deep image deblurring.
\newblock In {\em CVPR}, 2018.

\bibitem{vo2021attention}
Tu Vo.
\newblock Attention! stay focus!
\newblock In {\em CVPR Workshops}, 2021.

\bibitem{wadhwa2018synthetic}
Neal Wadhwa, Rahul Garg, David~E Jacobs, Bryan~E Feldman, Nori Kanazawa, Robert
  Carroll, Yair Movshovitz-Attias, Jonathan~T Barron, Yael Pritch, and Marc
  Levoy.
\newblock Synthetic depth-of-field with a single-camera mobile phone.
\newblock {\em ACM Transactions on Graphics}, 37(4):64, 2018.

\bibitem{yang2018seeing}
Jie Yang, Dong Gong, Lingqiao Liu, and Qinfeng Shi.
\newblock Seeing deeply and bidirectionally: A deep learning approach for
  single image reflection removal.
\newblock In {\em ECCV}, 2018.

\bibitem{yu2020bdd100k}
Fisher Yu, Haofeng Chen, Xin Wang, Wenqi Xian, Yingying Chen, Fangchen Liu,
  Vashisht Madhavan, and Trevor Darrell.
\newblock {BDD100K}: {A} diverse driving dataset for heterogeneous multitask
  learning.
\newblock In {\em CVPR}, 2020.

\bibitem{zhang2018single}
Xuaner Zhang, Ren Ng, and Qifeng Chen.
\newblock Single image reflection separation with perceptual losses.
\newblock In {\em CVPR}, 2018.

\bibitem{zhang20202}
Yinda Zhang, Neal Wadhwa, Sergio Orts-Escolano, Christian H{\"a}ne, Sean
  Fanello, and Rahul Garg.
\newblock Du2{N}et: Learning depth estimation from dual-cameras and
  dual-pixels.
\newblock In {\em ECCV}, 2020.

\bibitem{zhou2019spatio}
Shangchen Zhou, Jiawei Zhang, Jinshan Pan, Haozhe Xie, Wangmeng Zuo, and Jimmy
  Ren.
\newblock Spatio-temporal filter adaptive network for video deblurring.
\newblock In {\em ICCV}, 2019.

\end{thebibliography}
}

\end{document}